\documentclass{article}
\usepackage[a4paper, top=3cm, bottom=4cm, left=3.5cm, right=3.5cm]{geometry}
\usepackage[utf8]{inputenc}
\PassOptionsToPackage{table}{xcolor}
\usepackage{authblk}
\usepackage{hyperref}
\usepackage{multirow}
\usepackage{booktabs}

\usepackage{siunitx}
\usepackage{cleveref}

\usepackage{tikz}
\usetikzlibrary{external}
\usepackage{pgfplots}
\pgfplotsset{compat=1.9}
\usepgfplotslibrary{statistics}
\usetikzlibrary{arrows.meta,patterns}

\usepackage{graphicx}
\usepackage{float}
\usepackage{subcaption}
\graphicspath{{./figures/}}

\title{\textbf{Robot Gaze During Autonomous Navigation and its Effect on Social Presence}}
\author[1]{Kerry He}
\author[1]{Wesley P. Chan}
\author[2]{Akansel Cosgun}
\author[1]{Albin Joy}
\author[3]{Elizabeth A. Croft}
\affil[1]{Monash University, Australia}
\affil[2]{Deakin University, Australia}
\affil[3]{University of Victoria, Canada}
\date{}

\begin{document}

\maketitle


\abstract{As robots have become increasingly common in human-rich environments, it is critical that they are able to exhibit social cues to be perceived as a cooperative and socially-conformant team member. We investigate the effect of robot gaze cues on people's subjective perceptions of a mobile robot as a socially present entity in three common hallway navigation scenarios. The tested robot gaze behaviors were path-oriented (looking at its own future path), or person-oriented (looking at the nearest person), with fixed-gaze as the control. We conduct a real-world study with 36 participants who walked through the hallway, and an online study with 233 participants who were shown simulated videos of the same scenarios. Our results suggest that the preferred gaze behavior is scenario-dependent. Person-oriented gaze behaviors which acknowledge the presence of the human are generally preferred when the robot and human cross paths. However, this benefit is diminished in scenarios that involve less implicit interaction between the robot and the human.}



\section{Introduction}

As autonomous mobile robots become more capable of navigating indoors, they will start appearing more often in a wide variety of human-populated locations such as households, offices and public spaces. In addition to being able to safely maneuver around humans, robots operating in populated environments need to be equipped with appropriate social skills and behave in a socially acceptable manner~\cite{Kruse2013, lasota2017survey}. Humans typically navigate cooperatively when moving among other people. A moving robot should similarly be capable of developing a mutual understanding between itself and the humans in its vicinity when navigating among other people. This will allow both the robot and human to accurately interpret the actions of one another and allow for predictable movements~\cite{klein2005}. 

It is well known that humans use head and/or eye gaze (the direction in which one appears to be looking) during navigation and in other social situations to communicate intent and display mental states. Extensive research that has been conducted regarding interpersonal communication suggests that more than half of the meaning in social situations is communicated non-verbally~\cite{burgoon1994}. Therefore, it is critical that non-verbal cues such as the head gaze are integrated into existing navigation algorithms to improve the social aspects of the robot. 

To enable effective human-robot collaboration and teaming, the robot must be viewed as a socially capable agent by humans. In the existing Human-Robot Interaction (HRI) literature, this is referred to as the social presence of the robot, which is defined as the “perception of the [robot] with whom the [person] is interacting”~\cite{biocca2003}. With the goal of enabling social robots to operate more harmoniously and effectively around people, this paper investigates how different robot gaze behaviors during different navigation scenarios affect the robot's perceived social presence.


\section{Background}

In this section, we briefly review the existing literature on the use of mobile robots and social cues in HRI, followed by a deeper look at the use of gaze behaviors employed by robots for autonomous mobile robots.

\subsection{Mobile Robots in HRI}
As opposed to robots with a fixed base, mobile robots have access to a global workspace, and are therefore advantageous for tasks which require the robot to traverse between multiple areas. Fetch-and-carry tasks are a common application of mobile robots, typically contextualized as an assistive robot in household~\cite{srinivasa2010herb,cosgun2018context} or warehouse environments~\cite{bolu2021adaptive}. For these tasks, how to navigate in human-populated spaces is an important field of research. Trajectory planning in the presence of humans~\cite{dautenhahn2006may,sisbot2007human} considers aspects such as safety and visibility to the human, while also abiding by social conventions or human preferences. Proxemics~\cite{wiltshire2013,mead2017autonomous} considers the distance a robot should maintain from a human, which is important to maximizing the robot's perceived safety. In situations when the robot and human must give way to each other, effective communication and reactive planning is beneficial to cooperatively avoid collisions with each other~\cite{che2020efficient}. Another application of mobile robots in HRI is robot-to-human handovers, where the mobile nature of the robot allows it to perform fetch-and-carry tasks within a global workspace. Mainprice et al.~\cite{mainprice2012sharing} parameterizes a handover trajectory based on the effort that should be shared between the robot and human to perform a handover. He et al.~\cite{he2022go} studies mobile handovers where the base continues to move during physical transfer to improve efficiency of the handover.

\subsection{Social Cues in HRI}
Social cues are used by robots to communicate various types of information to nearby humans. Cid et al.~\cite{cid2014muecas} uses a robotic head with mechanisms to control neck, eye, eyebrow, and mouth motions to mimic human facial expressions. Similarly, Zecca et al. combines facial expressions with full body motions to recreate body language~\cite{zecca2009whole}. However these require complex mechanisms to recreate recognizable expressions. Semantic-free audio cues have also seen some interest in HRI, however is still relatively unexplored~\cite{yilmazyildiz2016review}. Tatarian et al.~\cite{tatarian2021does} investigated the use of multiple social cues simultaneously, including proxemics, gaze, gestures, and dialogue, and found that each social cue had a distinct effects on how the robot was perceived.

An important type of social cue is gaze cues~\cite{admoni2017social}. Gazes have been investigated in a variety of HRI contexts, for communicating different types of information. Moon et al.~\cite{moon2014meet} uses gaze cues during a robot-to-human handover to communicate handover location and timing information. Terziouglu et al.~\cite{terziouglu2020designing} utilize a variety of gaze cues, including gazes towards the task and target, as well as gazes towards the human collaborator to acknowledge completion of a task. In navigational scenarios, gaze can be used to convey navigational intent~\cite{hart2020using} or social presence with humans~\cite{khambhaita2016,wiltshire2013}. The latter is the primary purpose for which we employ gaze behaviors in our current study, while acknowledging that other interpretations of the gaze may be possible distractors from the true intention of the robot.

\subsection{Gaze in Navigational Scenarios}

Existing studies have investigated different aspects of robot head gaze during navigation, including different types of gaze behaviors and navigation scenarios. However, conflicting results have been reported regarding effects of gaze behaviors on the robot's perceived social presence \cite{khambhaita2016,wiltshire2013}. The primary difference between the two studies was the difference in the initiation timing of the gaze behavior and the navigation scenario explored. In the study conducted by Wiltshire et al.~\cite{wiltshire2013}, a robot and a person are engaged in a series of interactions in a hallway navigation scenario where the person must give way to the robot crossing paths with the person perpendicularly. During each of the interactions, nonverbal cues of the robot such as proxemics or gaze behaviors were altered. From their user study, it was concluded that altering the gaze behavior of the robot did not result in a higher social presence for the robot, and other non-verbal cues such as the proxemic behaviors exhibited by the robot are of more importance. In contrast, the study conducted by Khambhaita et al.~\cite{khambhaita2016}, which used very similar head gaze behaviors, concluded that altering the gaze behavior of the robot resulted in higher social presence. The gaze initiation timings between both of these studies were not the same, which may have contributed to the difference in the results. Furthermore, the navigation scenario used by Khambhaita et al. is not the same as what was used by Wiltshire et al. Therefore, it is still unclear and worthwhile to explore whether the appropriate gaze behavior, as well as the timings of gaze behavior execution, are dependent on the navigation scenario.

Another aspect of gaze during navigation is gaze fixation duration. Studies regarding gaze behaviors of adults during natural locomotion have found that they initiate gaze fixation \SI{1.72}{\second} before encountering an obstacle~\cite{hessels2020}. It was found that the duration of the fixation depended on the obstacle in question. For obstacles that were designed to be stimulating (stickers were placed on the obstacle), the fixation duration was found to be \SI{0.53}{\second}, while regular obstacles had a fixation duration of \SI{0.2}{\second}. However, the appropriate fixation duration for robots during navigation is still to be determined and is worth investigating.


\section{Research Questions}
As identified in the previous section, existing works on robot gaze during navigation has produced some conflicting results. In particular, whether and how navigation scenarios affect the perceived social presence of a robot along with robot gaze behavior has remained largely unexplored. Hence, our current work aims to address the following research questions: 
\begin{enumerate}
    \item How do different robot gaze behaviors during navigation affect people's perception of the robot as a socially present entity? 
    \item Does the appropriate gaze behavior vary depending on the navigation scenario? 
    \item How is the predictability of the robot affected by the use of different gaze behaviors?
\end{enumerate}

To answer these research questions, we conducted simulated and real-world user studies to investigate people's perceived social presence and predictability of a robot, when different gaze behaviors are used in different navigation scenarios.


\section{User Study Design}

\subsection{Independent Variables}

To answer our research questions, our study consists of a two-factor design, examining four gaze behaviors and three navigation scenarios, for a total of twelve conditions. These variables are detailed in the following sections.

\subsubsection{Gaze behaviors}
We tested four gaze behaviors in our study, visualized in Fig.~\ref{fig:gazes}: 

\begin{figure}[h]
    \centering
    \input{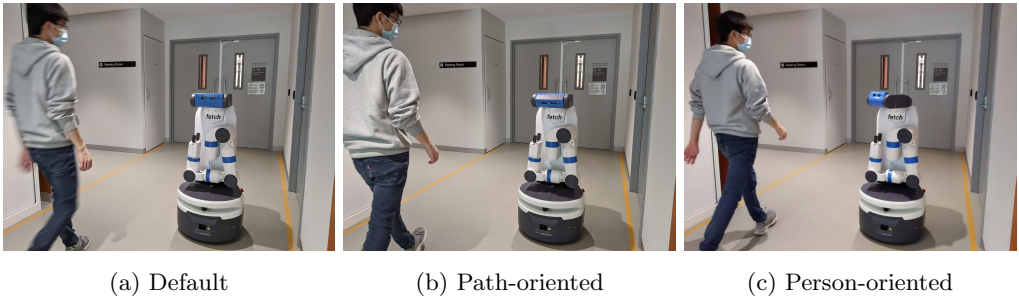}
    \caption{Different gaze behaviors which were employed for the user study.}
    \label{fig:gazes}
\end{figure}

\begin{itemize}
    \item \textbf{Default (D)}: The robot always looks forwards, and does not move its head. This gaze serves as the control for the user study.
    \item \textbf{Path Oriented (PO)}: The robot looks at a point on the ground \SI{1.5}{\meter} ahead along the path it is planning to travel in.
    \item \textbf{Person-Oriented Short (POS)} and \textbf{Long (POL)}: The robot ``acknowledges" the person by briefly directing its gaze at the person when passing them. Two gaze fixation timings are tested based on the timings used by humans during locomotion~\cite{hessels2020}. These gaze periods are \SI{0.2}{\second} for POS, and \SI{0.53}{\second} for POL. The robot uses the default gaze before and after it looks at the human.
\end{itemize}

\subsubsection{Navigational Scenarios}
\begin{figure}[h]
    \centering
    \tikzset{every picture/.style={line width=0.75pt}} 

\begin{tikzpicture}[x=0.75pt,y=0.75pt,yscale=-1,xscale=1]

\draw    (0,23) -- (0,130) ;
\draw    (80,23) -- (80,130) ;
\draw [color={rgb, 255:red, 255; green, 0; blue, 0 }  ,draw opacity=1 ] [dash pattern={on 4.5pt off 4.5pt}]  (20,26) -- (20,130) ;
\draw [shift={(20,130)}, rotate = 90] [color={rgb, 255:red, 255; green, 0; blue, 0 }  ,draw opacity=1 ][fill={rgb, 255:red, 255; green, 0; blue, 0 }  ,fill opacity=1 ][line width=0.75]      (0, 0) circle [x radius= 3.35, y radius= 3.35]   ;
\draw [shift={(20,23)}, rotate = 90] [fill={rgb, 255:red, 255; green, 0; blue, 0 }  ,fill opacity=1 ][line width=0.08]  [draw opacity=0] (10.72,-5.15) -- (0,0) -- (10.72,5.15) -- (7.12,0) -- cycle    ;
\draw  [dash pattern={on 4.5pt off 4.5pt}]  (60,127) -- (60,72.93) ;
\draw [shift={(60,130)}, rotate = 270] [fill={rgb, 255:red, 0; green, 0; blue, 0 }  ][line width=0.08]  [draw opacity=0] (10.72,-5.15) -- (0,0) -- (10.72,5.15) -- (7.12,0) -- cycle    ;
\draw  [draw opacity=0][dash pattern={on 4.5pt off 4.5pt}] (60,70) .. controls (60,70) and (60,70) .. (60,70) .. controls (48.95,70) and (40,61.08) .. (40,50.07) .. controls (40,39.06) and (48.95,30.13) .. (60,30.13) -- (60,50.07) -- cycle ; \draw  [dash pattern={on 4.5pt off 4.5pt}] (60,70) .. controls (60,70) and (60,70) .. (60,70) .. controls (48.95,70) and (40,61.08) .. (40,50.07) .. controls (40,39.06) and (48.95,30.13) .. (60,30.13) ;  
\draw  [dash pattern={on 4.5pt off 4.5pt}]  (60,30.13) -- (60,23) ;
\draw [shift={(60,23)}, rotate = 270] [color={rgb, 255:red, 0; green, 0; blue, 0 }  ][fill={rgb, 255:red, 0; green, 0; blue, 0 }  ][line width=0.75]      (0, 0) circle [x radius= 3.35, y radius= 3.35]   ;
\draw  [fill={rgb, 255:red, 207; green, 207; blue, 207 }  ,fill opacity=1 ] (55,40) -- (75,40) -- (75,60) -- (55,60) -- cycle ;
\draw    (150,101.47) -- (150,130) ;
\draw    (230,23) -- (230,130) ;
\draw [color={rgb, 255:red, 255; green, 0; blue, 0 }  ,draw opacity=1 ] [dash pattern={on 4.5pt off 4.5pt}]  (170,26) -- (170,130) ;
\draw [shift={(170,130)}, rotate = 90] [color={rgb, 255:red, 255; green, 0; blue, 0 }  ,draw opacity=1 ][fill={rgb, 255:red, 255; green, 0; blue, 0 }  ,fill opacity=1 ][line width=0.75]      (0, 0) circle [x radius= 3.35, y radius= 3.35]   ;
\draw [shift={(170,23)}, rotate = 90] [fill={rgb, 255:red, 255; green, 0; blue, 0 }  ,fill opacity=1 ][line width=0.08]  [draw opacity=0] (10.72,-5.15) -- (0,0) -- (10.72,5.15) -- (7.12,0) -- cycle    ;
\draw  [dash pattern={on 4.5pt off 4.5pt}]  (143,87.2) -- (210,87.2) -- (210,23) ;
\draw [shift={(210,23)}, rotate = 270] [color={rgb, 255:red, 0; green, 0; blue, 0 }  ][fill={rgb, 255:red, 0; green, 0; blue, 0 }  ][line width=0.75]      (0, 0) circle [x radius= 3.35, y radius= 3.35]   ;
\draw [shift={(140,87.2)}, rotate = 0] [fill={rgb, 255:red, 0; green, 0; blue, 0 }  ][line width=0.08]  [draw opacity=0] (10.72,-5.15) -- (0,0) -- (10.72,5.15) -- (7.12,0) -- cycle    ;
\draw    (150,23) -- (150,72.93) ;
\draw    (300,101.47) -- (300,130) ;
\draw    (380,23) -- (380,130) ;
\draw [color={rgb, 255:red, 255; green, 0; blue, 0 }  ,draw opacity=1 ] [dash pattern={on 4.5pt off 4.5pt}]  (320,26) -- (320,130) ;
\draw [shift={(320,130)}, rotate = 90] [color={rgb, 255:red, 255; green, 0; blue, 0 }  ,draw opacity=1 ][fill={rgb, 255:red, 255; green, 0; blue, 0 }  ,fill opacity=1 ][line width=0.75]      (0, 0) circle [x radius= 3.35, y radius= 3.35]   ;
\draw [shift={(320,23)}, rotate = 90] [fill={rgb, 255:red, 255; green, 0; blue, 0 }  ,fill opacity=1 ][line width=0.08]  [draw opacity=0] (10.72,-5.15) -- (0,0) -- (10.72,5.15) -- (7.12,0) -- cycle    ;
\draw  [dash pattern={on 4.5pt off 4.5pt}]  (360,127) -- (360,87.2) -- (290,87.2) ;
\draw [shift={(290,87.2)}, rotate = 180] [color={rgb, 255:red, 0; green, 0; blue, 0 }  ][fill={rgb, 255:red, 0; green, 0; blue, 0 }  ][line width=0.75]      (0, 0) circle [x radius= 3.35, y radius= 3.35]   ;
\draw [shift={(360,130)}, rotate = 270] [fill={rgb, 255:red, 0; green, 0; blue, 0 }  ][line width=0.08]  [draw opacity=0] (10.72,-5.15) -- (0,0) -- (10.72,5.15) -- (7.12,0) -- cycle    ;
\draw    (300,23) -- (300,72.93) ;

\draw (20.5,11.5) node  [font=\footnotesize,color={rgb, 255:red, 255; green, 0; blue, 0 }  ,opacity=1 ] [align=left] {Human};
\draw (170.5,11.5) node  [font=\footnotesize,color={rgb, 255:red, 255; green, 0; blue, 0 }  ,opacity=1 ] [align=left] {Human};
\draw (320.5,11.5) node  [font=\footnotesize,color={rgb, 255:red, 255; green, 0; blue, 0 }  ,opacity=1 ] [align=left] {Human};
\draw (60.5,11.5) node  [font=\footnotesize] [align=left] {Robot};
\draw (210,11.5) node  [font=\footnotesize] [align=left] {Robot};
\draw (360,11.5) node  [font=\footnotesize] [align=left] {Robot};
\draw (40.5,149.5) node  [font=\small,color={rgb, 255:red, 0; green, 0; blue, 0 }  ,opacity=1 ] [align=left] {(a) Two way};
\draw (190,149.5) node  [font=\small,color={rgb, 255:red, 0; green, 0; blue, 0 }  ,opacity=1 ] [align=left] {(b) Robot enters room};
\draw (340.5,149.5) node  [font=\small,color={rgb, 255:red, 0; green, 0; blue, 0 }  ,opacity=1 ] [align=left] {(c) Robot exits room};

\end{tikzpicture}
    \caption{Illustrative layout of the navigation scenarios considered for the user study.}
    \label{fig:scenario}
\end{figure}
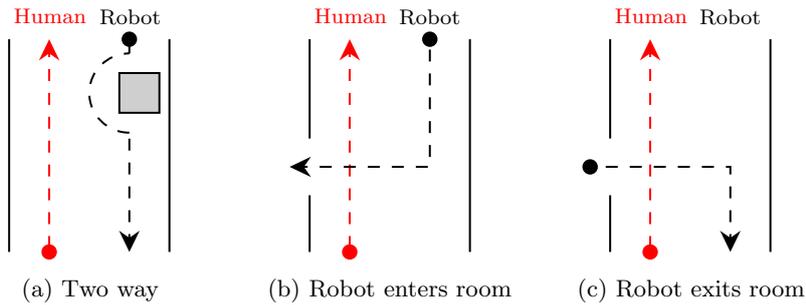

Our study incorporates three different navigation scenarios as illustrated in Fig.~\ref{fig:scenario}. Hallway scenarios were chosen as they are typical in many buildings and are the most common scenario used in existing literature regarding gaze behaviors. Furthermore, these scenarios permit us to make direct comparisons with existing works.

\begin{itemize}
    \item \textbf{Two way (TW)}: The person and robot navigate down a hallway starting from opposite ends. An obstacle is placed in the path of the robot to accentuate the difference between the default and path-oriented gaze behaviors.
    
    \item \textbf{Robot enters room (ENT)}: The robot moves down the hallway then enters a room on the human's left. The robot briefly pauses before entering to attempt to give way to the human but always decides to enter the room first as the human is too far away to give way to.

    \item \textbf{Robot exits room (EXT)}: The robot begins inside a room to the human's left, where it is not initially visible to the human. The robot exits the room before turning right and moving down the hallway. The person is forced to give way to the robot for it to make its turn.
\end{itemize}

\subsection{Dependent Variables}

To measure the social presence of the robot, we use the Social Presence Inventory (SPI) developed by Harms and Biocca~\cite{Harms2004}. Specifically, we use the following dimensions:

\begin{itemize}
    \item \textbf{Co-Presence (CP)}: The SPI defines co-presence as ``the degree to which the observer believes he/she is not alone and secluded, their level of peripheral or focal awareness of the other, and their sense of the degree to which the other is peripherally or focally aware of them''. Co-presence fundamentally measures subjective perception rather than the objective condition of being observable. This means that co-presence is well suited for self-report measures such as a Likert scale that indicates the level of awareness of each other. The sensory awareness of agents in a given interaction can also be measured using observations such as measuring eye fixation, proxemic behavior exhibited, or physiological responses. The primary challenges with such methods are the difficulty in data collection, especially with a remote user study.
    
    \item \textbf{Perceived Message Understanding (PMU)}: The SPI defines PMU as ``the ability of the user to understand the message being received from the interactant as well as their perception of the interactant’s level of message understanding''. These measures can be used to evaluate how the legibility of the robot changes when gaze behavior is altered. 
    
    \item \textbf{Perceived Behavioral Interdependence (PBI)}: The SPI defines PBI as the ``extent to which a user’s behavior affects and is affected by the interactant’s behavior''. In a given interaction, the key sense of access to others is based on the degree to which the agents appear to interact with each other. This can include explicit behaviors such as waving of hands to acknowledge the presence of each other or non-explicit behaviors such as eye contact.
\end{itemize}

In addition to the social presence metrics, we also measure the following factors:

\begin{itemize}
    \item \textbf{Predictability}: The predictability of the motion of the robot. This is different from the PMU of the robot as the participant can potentially understand what task the robot is trying to complete, but is not sure about how it is going to do it.
    \item \textbf{Perceived Safety}: How safe the person feels in the presence of the robot.
    \item \textbf{Naturalness}: How similar the robot behaves compared to the typical human.
\end{itemize}

All dependent variables are quantitative measures reflecting the subjective assessment of the human. These are acquired in the form of post-experiment survey questions, and measured using a Likert-scale.

\subsection{Hypotheses}
We expect human-like gazes, such as person-oriented gazes which establish eye-contact with a human when they acknowledge the human's presence, to feel more socially conformant and socially present (Hypothesis 1). As a result, we also expect person-oriented gazes to feel safer and more natural (Hypothesis 3). However, people may interpret the robot's gaze direction as the direction the robot will navigate towards. Therefore, in scenarios where the human is not in the same direction that the robot will navigate towards, such as the ENT scenario, people may be surprised when the robot moves in one direction after gazing in a different direction (Hypothesis 2). Therefore, we make the following hypotheses:

\begin{itemize}
    \item \textbf{H1}: Person-oriented gazes will have higher perceived social presence compared to other gazes in all scenarios.

    \item \textbf{H2}: Person-oriented gazes will be perceived as less predictable for the ENT scenario.

    \item \textbf{H3}: Person-oriented gazes will improve the perceived safety and naturalness of the robot for all scenarios.
\end{itemize}

\subsection{Implementation}

\begin{figure}[t]
    \centering
    \subfloat{{\includegraphics[width=5cm, angle=-90]{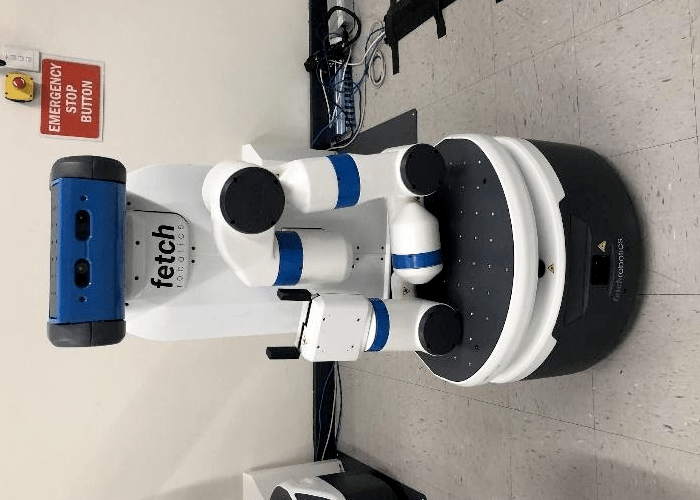} }}%
    \quad
    \subfloat{{\includegraphics[height=5cm]{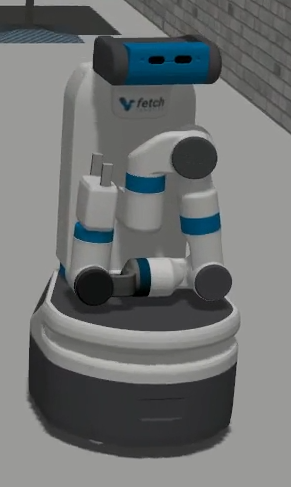} }}%
    \caption{Real-world (left) and simulated (right) Fetch Mobile Manipulator}
    \label{fig:fetch}
\end{figure}

\subsubsection{Robotic Platform}
The Fetch Mobile Manipulator robot, shown in Fig.~\ref{fig:fetch}, was chosen as the research platform as it has a head module that is capable of both pan and tilt, as well as a nonholonomic mobile base which can maneuver around a global workspace. 

\subsubsection{Simulation Environment}
For the simulated study, the Gazebo simulator\footnote{\url{http://gazebosim.org/}} was used to generate videos of the Fetch robot performing each gaze and scenario.

\subsubsection{Real-world Implementation}
\begin{figure}[t]
    \centering
    \tikzset{every picture/.style={line width=0.75pt}} 

\begin{tikzpicture}[x=0.75pt,y=0.75pt,yscale=-1,xscale=1]

\draw  [fill={rgb, 255:red, 200; green, 255; blue, 200 }  ,fill opacity=1 ] (332.58,392.69) -- (404.21,392.69) -- (404.21,421.35) -- (332.58,421.35) -- cycle ;
\draw  [fill={rgb, 255:red, 255; green, 200; blue, 200 }  ,fill opacity=1 ] (432.87,392.69) -- (504.51,392.69) -- (504.51,421.35) -- (432.87,421.35) -- cycle ;
\draw  [fill={rgb, 255:red, 225; green, 225; blue, 225 }  ,fill opacity=1 ] (10.21,364.04) -- (81.85,364.04) -- (81.85,392.69) -- (10.21,392.69) -- cycle ;
\draw  [fill={rgb, 255:red, 200; green, 255; blue, 200 }  ,fill opacity=1 ] (146.32,364.04) -- (217.96,364.04) -- (217.96,392.69) -- (146.32,392.69) -- cycle ;
\draw    (404.21,407.02) -- (429.87,407.02) ;
\draw [shift={(432.87,407.02)}, rotate = 180] [fill={rgb, 255:red, 0; green, 0; blue, 0 }  ][line width=0.08]  [draw opacity=0] (10.72,-5.15) -- (0,0) -- (10.72,5.15) -- (7.12,0) -- cycle    ;
\draw    (81.85,378.36) -- (143.32,378.36) ;
\draw [shift={(146.32,378.36)}, rotate = 180] [fill={rgb, 255:red, 0; green, 0; blue, 0 }  ][line width=0.08]  [draw opacity=0] (10.72,-5.15) -- (0,0) -- (10.72,5.15) -- (7.12,0) -- cycle    ;
\draw    (217.96,378.36) -- (303.92,378.36) -- (303.92,407.02) -- (329.58,407.02) ;
\draw [shift={(332.58,407.02)}, rotate = 180] [fill={rgb, 255:red, 0; green, 0; blue, 0 }  ][line width=0.08]  [draw opacity=0] (10.72,-5.15) -- (0,0) -- (10.72,5.15) -- (7.12,0) -- cycle    ;
\draw  [fill={rgb, 255:red, 200; green, 255; blue, 200 }  ,fill opacity=1 ] (146.32,421.35) -- (217.96,421.35) -- (217.96,450) -- (146.32,450) -- cycle ;
\draw    (217.96,435.67) -- (303.92,435.67) -- (303.92,407.02) -- (329.58,407.02) ;
\draw [shift={(332.58,407.02)}, rotate = 180] [fill={rgb, 255:red, 0; green, 0; blue, 0 }  ][line width=0.08]  [draw opacity=0] (10.72,-5.15) -- (0,0) -- (10.72,5.15) -- (7.12,0) -- cycle    ;

\draw  [draw opacity=0]  (328.89,388.02) -- (407.89,388.02) -- (407.89,426.02) -- (328.89,426.02) -- cycle  ;
\draw (368.39,407.02) node  [font=\footnotesize] [align=left] {\begin{minipage}[lt]{50.77pt}\setlength\topsep{0pt}
\begin{center}
Head Control\\Node
\end{center}

\end{minipage}};
\draw  [draw opacity=0]  (431.19,388.02) -- (506.19,388.02) -- (506.19,426.02) -- (431.19,426.02) -- cycle  ;
\draw (468.69,407.02) node  [font=\footnotesize] [align=left] {\begin{minipage}[lt]{48.51pt}\setlength\topsep{0pt}
\begin{center}
Point-Head\\Action~Client
\end{center}

\end{minipage}};
\draw  [draw opacity=0]  (13.53,367.86) -- (78.53,367.86) -- (78.53,388.86) -- (13.53,388.86) -- cycle  ;
\draw (46.03,378.36) node  [font=\footnotesize] [align=left] {\begin{minipage}[lt]{41.25pt}\setlength\topsep{0pt}
\begin{center}
Navigation
\end{center}

\end{minipage}};
\draw  [draw opacity=0]  (148.14,367.86) -- (216.14,367.86) -- (216.14,388.86) -- (148.14,388.86) -- cycle  ;
\draw (182.14,378.36) node  [font=\footnotesize] [align=left] {\begin{minipage}[lt]{43.52pt}\setlength\topsep{0pt}
\begin{center}
Move Base
\end{center}

\end{minipage}};
\draw  [draw opacity=0]  (143.64,425.17) -- (220.64,425.17) -- (220.64,446.17) -- (143.64,446.17) -- cycle  ;
\draw (182.14,435.67) node  [font=\footnotesize] [align=left] {\begin{minipage}[lt]{49.41pt}\setlength\topsep{0pt}
\begin{center}
Leg Detector
\end{center}

\end{minipage}};
\draw  [draw opacity=0]  (81.07,360.34) -- (142.07,360.34) -- (142.07,381.34) -- (81.07,381.34) -- cycle  ;
\draw (111.57,370.84) node  [font=\footnotesize] [align=left] {\begin{minipage}[lt]{38.53pt}\setlength\topsep{0pt}
\begin{center}
Odometry
\end{center}

\end{minipage}};
\draw  [draw opacity=0]  (212.94,418.01) -- (308.94,418.01) -- (308.94,439.01) -- (212.94,439.01) -- cycle  ;
\draw (260.94,428.51) node  [font=\footnotesize] [align=left] {\begin{minipage}[lt]{62.56pt}\setlength\topsep{0pt}
\begin{center}
Detected person
\end{center}

\end{minipage}};
\draw  [draw opacity=0]  (221.44,360.7) -- (300.44,360.7) -- (300.44,381.7) -- (221.44,381.7) -- cycle  ;
\draw (260.94,371.2) node  [font=\footnotesize] [align=left] {\begin{minipage}[lt]{50.77pt}\setlength\topsep{0pt}
\begin{center}
Planned path
\end{center}

\end{minipage}};

\end{tikzpicture}
    \caption{Head Control Node interacting with the existing ROS packages}
    \label{fig:system_diagram}
\end{figure}
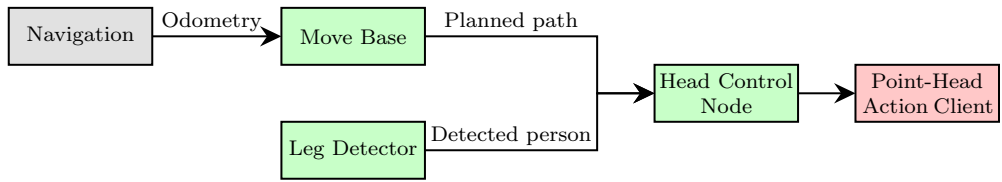

Our real-world implementation is summarized in the system diagram shown in Fig.~\ref{fig:system_diagram}. We implemented our system using the Robot Operating System (ROS)\footnote{\url{https://www.ros.org/}} framework. The Head Control Node commands the robot head motion to exhibit different gaze behaviors. We use the Leg Detector ROS package\footnote{\url{http://wiki.ros.org/leg_detector}} to detect and track people around the robot using the robot's laser scanner sensor, and the move\_base package from the ROS Navigation stack for planning robot base motion. The Head Control Node receives the positions of the detected person from the Leg Detector, and the planned path of the robot from the Move Base node. The gaze behavior is executed by sending motion commands to the Point-Head Action Client, which controls the pan-tilt motions of the robot's head. 

To avoid collisions with obstacles or people, a safety feature was implemented  where the robot would stop moving if the laser scanner detects an obstacle within a fixed distance (0.6m in our experiments).

\subsection{User Study Procedure}

\begin{figure}[t]
    \centering
    \includegraphics[scale = 0.3]{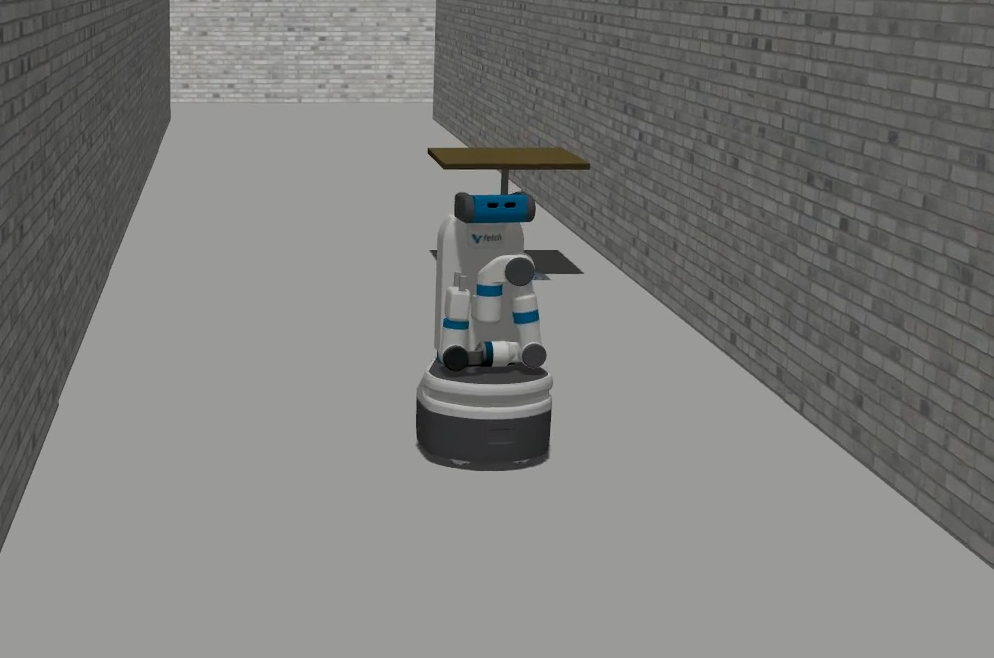}
    \caption{Snapshot of the simulation video used for the TW scenario}%
    \label{fig:two_way_snap}
\end{figure}

\subsubsection{Simulated video study}
Twelve videos\footnote{\url{https://tinyurl.com/mape3d2u}} addressing each condition were generated using the Gazebo simulator. Each participant was shown the videos in a randomized order, and participants were instructed to view the videos full screen. After watching each video, participants were given a survey asking them to rate a series of statements (shown in Table~\ref{table:survey}) on a 3-point Likert scale. We distributed our study through Amazon Mechanical Turk and we compensated each participant who successfully completed the study with \$2 USD. To ensure data quality, we included a control question asking the participant to identify which direction the robot went (left/right) at the end of the video. Data from those who failed the control question, or were incomplete, were rejected. 15 responses were rejected based on these criteria. In total we accepted responses from 233 participants (Male=142, Female=91).

\begin{table}[]
\caption{Statements which participants responded to for both the simulated and real-world user studies.}
\small
\centering
\setlength{\aboverulesep}{0pt}
\setlength{\belowrulesep}{0pt}
\setlength{\extrarowheight}{.75ex}
\begin{tabular}{llcc}
\toprule
\textbf{Outcome Variable} & \hspace{0.5cm} & \textbf{Statement} & \hspace{0.25cm} \\ \midrule
\rowcolor{black!5} 
\cellcolor{black!5}  & & You were noticed by the robot. & \\
\rowcolor{black!5}  
\cellcolor{black!5}  & & The robot caught your attention. & \\
\rowcolor{black!5}  
\multirow{-3}{*}{\cellcolor{black!5} \textbf{Co-Presence}} & & Your presence was obvious to the robot. & \\
\textbf{\begin{tabular}[c]{@{}l@{}}Perceived Message\\Understanding\end{tabular}} & & It was easy to understand the robot’s intentions. & \\
\rowcolor{black!5}  
\textbf{\begin{tabular}[c]{@{}l@{}}Perceived Behavioral\\ Interdependence\end{tabular}} & & The robot’s behavior was in direct response to your actions. & \\
\textbf{Predictability} & & The actions of the robot was predictable. & \\
\rowcolor{black!5}  
\textbf{Perceived Safety} & & You felt safe in the presence of the robot. & \\
\textbf{Naturalness} & & The actions of the robot seemed natural. & \\ \bottomrule
\end{tabular}
\label{table:survey}
\end{table}

\subsubsection{Real-world studies}

At the beginning of each experiment, the participant is shown simulated videos (the same as those used for the simulated studies) of the D gaze behavior for each of the three navigation scenarios so that they understand how they should behave for each scenario. Participants then experienced each of the 12 total gaze behavior and navigation scenario combinations. These were ran in a pseudo-randomized order using a Balanced Latin Square\footnote{\url{https://cs.uwaterloo.ca/~dmasson/tools/latin_square/}} to minimize effects caused by the specific ordering of trials in the experiment. Participants were informed about which navigation scenario they would face next, but not which gaze behavior the robot would exhibit. After each trial, participants were asked to respond to the same set of statements as the simulated study (shown in Table~\ref{table:survey}) using a 5-point Likert scale. Trials where either participants walked too quickly or the human detection algorithm failed were repeated. At the end of the experiment, participants were asked if they had any additional comments. Overall, we recruited 36 participants (Male=27, Female=9; 35 between ages 18-25, 1 aged 31-40) from the lab and University premises to participate in the real-world user studies\footnote{No external participants could be recruited for the real-world experiments due to the university COVID-19 policy.}. 


\section{Results}

\begin{figure*}[t]
\centering
\begin{tikzpicture}

    \definecolor{a}{RGB}{217,217,217}
    \definecolor{b}{RGB}{255,255,179}
    \definecolor{c}{RGB}{253,180,98}
    \definecolor{d}{RGB}{251,128,114}

    \pgfplotsset{
      /pgfplots/bar  cycle  list/.style={/pgfplots/cycle  list={%
            {a!67!black,fill=a,mark=none},%
            {b!67!black,fill=b,mark=none},%
            {c!67!black,fill=c,mark=none},%
            {d!67!black,fill=d,mark=none},%
         }
      },
    }

    \begin{axis}[
        name=mainplot,
        ybar,
        enlarge x limits=0.25,
        ylabel={Nat. (Simulated)},
        symbolic x coords={TW,ENT,EXT},
        xtick=data,
        xticklabels={,,},
        bar width=10pt,
        xtick pos=left,
        width = 0.55\columnwidth,
        height = 0.3\columnwidth,
        ymin = 1, ymax = 3.25,
        title=\small(a),
        title style={yshift=-1.25ex,},        
    ]
    
    \addplot+ [
        error bars/.cd,
            y dir=both, y explicit,
    ] coordinates {
        (TW,2.5489)  +- (0, 0.03996)
        (ENT,2.1319) +- (0, 0.04640)
        (EXT,2.2426) +- (0, 0.04645)
    };

    \addplot+ [
        error bars/.cd,
            y dir=both, y explicit,
    ] coordinates {
        (TW,2.4979)  +- (0, 0.03917)
        (ENT,2.2255) +- (0, 0.04720)
        (EXT,2.4043) +- (0, 0.04225)
    };
    
    \addplot+ [
        error bars/.cd,
            y dir=both, y explicit,
    ] coordinates {
        (TW,2.4894)  +- (0, 0.04142)
        (ENT,2.4426) +- (0, 0.04297)
        (EXT,2.5532) +- (0, 0.03712)
    };    
    
    \addplot+ [
        error bars/.cd,
            y dir=both, y explicit,
    ] coordinates {
        (TW,2.3872)  +- (0, 0.04335)
        (ENT,2.2851) +- (0, 0.04541)
        (EXT,2.3617) +- (0, 0.04345)
    };
    
    \draw [arrows={Bar[left]-Bar[right,width=17.5pt]}] 
        (axis cs:TW,2.725) ++(-18pt, 0) -- node[midway, above, yshift=-6.5pt]{*} ++(36pt,0);
        
    \draw [arrows={Bar[left,width=20pt,length=12pt]-Bar[right]}] 
        (axis cs:ENT,2.625) ++(-18pt, 0) -- node[midway, above, yshift=-6.5pt,xshift=3pt]{*} ++(24pt,0);
        
    \draw [arrows={Bar[left,width=15pt]-Bar[right]}] 
        (axis cs:EXT,2.575) ++(-18pt, 0) -- node[midway, above, yshift=-6.5pt]{*} ++(12pt,0);
        
    \draw [arrows={Bar[left,width=12.5pt]-Bar[right]}] 
        (axis cs:EXT,2.825) ++(-18pt, 0) -- node[midway, above, yshift=-6.5pt]{**} ++(24pt,0);
        
    \draw [arrows={Bar[left]-Bar[right,width=15pt]}] 
        (axis cs:EXT,2.7) ++(6pt, 0) -- node[midway, above, yshift=-6.5pt]{*} ++(12pt,0);

    \end{axis}
    \definecolor{a}{RGB}{204,235,197}
    \definecolor{b}{RGB}{128,177,211}
    \definecolor{c}{RGB}{188,128,189}    
    \begin{axis}[
        name=secondplot,
        at={(mainplot.north east)},
        xshift=0.5cm,
        anchor=north west,    
        ybar,
        enlarge x limits=0.175,
        symbolic x coords={D,PO,POS,POL},
        xtick=data,
        xticklabels={,,},
        yticklabels={,,},
        bar width=10pt,
        xtick pos=left,
        width = 0.55\columnwidth,
        height = 0.3\columnwidth,
        ymin = 1, ymax = 3.25,
        title=\small(b),
        title style={yshift=-1.25ex,},        
    ]
    
    \addplot+ [
        error bars/.cd,
            y dir=both, y explicit,
    ] coordinates {
        (D,2.5489)  +- (0, 0.03996)
        (PO,2.4979)  +- (0, 0.03917)
        (POS,2.4894)  +- (0, 0.04142)
        (POL,2.3872)  +- (0, 0.04335)
    };

    \addplot+ [
        error bars/.cd,
            y dir=both, y explicit,
    ] coordinates {
        (D,2.1319) +- (0, 0.04640)
        (PO,2.2255) +- (0, 0.04720)
        (POS,2.4426) +- (0, 0.04297)
        (POL,2.2851) +- (0, 0.04541)
    };
    
    \addplot+ [
        error bars/.cd,
            y dir=both, y explicit,
    ] coordinates {
        (D,2.2426) +- (0, 0.04645)
        (PO,2.4043) +- (0, 0.04225)
        (POS,2.5532) +- (0, 0.03712)
        (POL,2.3617) +- (0, 0.04345)
    };

    \draw [arrows={Bar[left]-Bar[right,width=22.5pt,length=12pt]}] 
        (axis cs:D,2.725) ++(-12pt, 0) -- node[midway, above, yshift=-6.5pt,xshift=-3pt]{**} ++(24pt,0);
        
    \draw [arrows={Bar[left,width=7.5pt]-Bar[right]}] 
        (axis cs:PO,2.7) ++(-12pt, 0) -- node[midway, above, yshift=-6.5pt]{**} ++(12pt,0);

    \draw [arrows={Bar[left,width=15pt]-Bar[right]}] 
        (axis cs:PO,2.575) ++(0pt, 0) -- node[midway, above, yshift=-6.5pt]{*} ++(12pt,0);        

    \end{axis}
    \definecolor{a}{RGB}{217,217,217}
    \definecolor{b}{RGB}{255,255,179}
    \definecolor{c}{RGB}{253,180,98}
    \definecolor{d}{RGB}{251,128,114}    
    \begin{axis}[
        at={(mainplot.below south west)},
        yshift=-0.35cm,
        anchor=north west,
        ybar,
        enlarge x limits=0.25,
        legend style={at={(0.5,-0.25)},
        anchor=north,legend columns=-1,
        /tikz/every even column/.append style={column sep=0.5cm}},
        ylabel={Nat. (Real-world)},
        symbolic x coords={TW,ENT,EXT},
        xtick=data,
        bar width=10pt,
        xtick pos=left,
        width = 0.55\columnwidth,
        height = 0.3\columnwidth,
        ymin = 1, ymax = 5.49,
        legend image code/.code={\draw [#1] (0cm,-0.075cm) rectangle (0.2cm,0.125cm); },        title=\small(c),
        title style={yshift=-1.25ex,},
    ]
    
    \addplot+ [
        error bars/.cd,
            y dir=both, y explicit,
    ] coordinates {
        (TW,3.6111)  +- (0, 0.1794)
        (ENT,3.4167) +- (0, 0.1842)
        (EXT,3.1667) +- (0, 0.1890)
    };
    
    \addplot+ [
        error bars/.cd,
            y dir=both, y explicit,
    ] coordinates {
        (TW,3.3889)  +- (0, 0.2003)
        (ENT,3.6944) +- (0, 0.1985)
        (EXT,3.1944) +- (0, 0.1728)
    };    
    
    \addplot+ [
        error bars/.cd,
            y dir=both, y explicit,
    ] coordinates {
        (TW,3.6389)  +- (0, 0.1917)
        (ENT,3.7778) +- (0, 0.1497)
        (EXT,3.9722) +- (0, 0.1516)
    };
    
    \addplot+ [
        error bars/.cd,
            y dir=both, y explicit,
    ] coordinates {
        (TW,3.8333)  +- (0, 0.1804)
        (ENT,3.8333) +- (0, 0.1618)
        (EXT,3.9722) +- (0, 0.1516)
    };    
      
    \legend{D,PO,POS,POL}
    
    \draw [arrows={Bar[left,width=32.5pt]-Bar[right,width=7.5pt,length=12pt]}] 
        (axis cs:EXT,4.55) ++(-6pt, 0) -- node[midway, above, yshift=-6.5pt,xshift=-3pt]{**} ++(24pt,0);
        
    \draw [arrows={Bar[left,width=52.5pt]-Bar[right,width=7.5pt,length=12pt]}] 
        (axis cs:EXT,5.1) ++(-18pt, 0) -- node[midway, above, yshift=-6.5pt,xshift=-3pt]{*} ++(36pt,0);        
    
    \end{axis}
    \definecolor{a}{RGB}{204,235,197}
    \definecolor{b}{RGB}{128,177,211}
    \definecolor{c}{RGB}{188,128,189}    
    \begin{axis}[
        at={(secondplot.below south west)},
        yshift=-0.35cm,
        anchor=north west,  
        ybar,
        enlarge x limits=0.175,
        legend style={at={(0.5,-0.25)},
        anchor=north,legend columns=-1,
        /tikz/every even column/.append style={column sep=0.5cm}},
        symbolic x coords={D,PO,POS,POL},
        xtick=data,
        yticklabels={,,},
        bar width=10pt,
        xtick pos=left,
        width = 0.55\columnwidth,
        height = 0.3\columnwidth,
        ymin = 1, ymax = 5.49,
        legend image code/.code={\draw [#1] (0cm,-0.075cm) rectangle (0.2cm,0.125cm); },
        title=\small(d),
        title style={yshift=-1.25ex,},        
    ]
    
    \addplot+ [
        error bars/.cd,
            y dir=both, y explicit,
    ] coordinates {
        (D,3.6111)  +- (0, 0.1794)
        (PO,3.3889)  +- (0, 0.2003)
        (POS,3.6389)  +- (0, 0.1917)
        (POL,3.8333)  +- (0, 0.1804)
    };

    \addplot+ [
        error bars/.cd,
            y dir=both, y explicit,
    ] coordinates {
        (D,3.4167) +- (0, 0.1842)
        (PO,3.6944) +- (0, 0.1985)
        (POS,3.7778) +- (0, 0.1497)
        (POL,3.8333) +- (0, 0.1618)
    };
    
    \addplot+ [
        error bars/.cd,
            y dir=both, y explicit,
    ] coordinates {
        (D,3.1667) +- (0, 0.1890)
        (PO,3.1944) +- (0, 0.1728)
        (POS,3.9722) +- (0, 0.1516)
        (POL,3.9722) +- (0, 0.1516)
    };
    
    \legend{TW,ENT,EXT}
    
        
    \end{axis}        
\end{tikzpicture}
\caption{Results obtained for Naturalness for simulated (top) and real-world (bottom) user studies. Results are shown grouped by gaze behavior (left) and navigation scenario (right). Significant paired differences are indicated with ($*$) for $p<0.05$, and with ($**$) for $p<0.001$.}
\label{fig:nat}
\end{figure*}
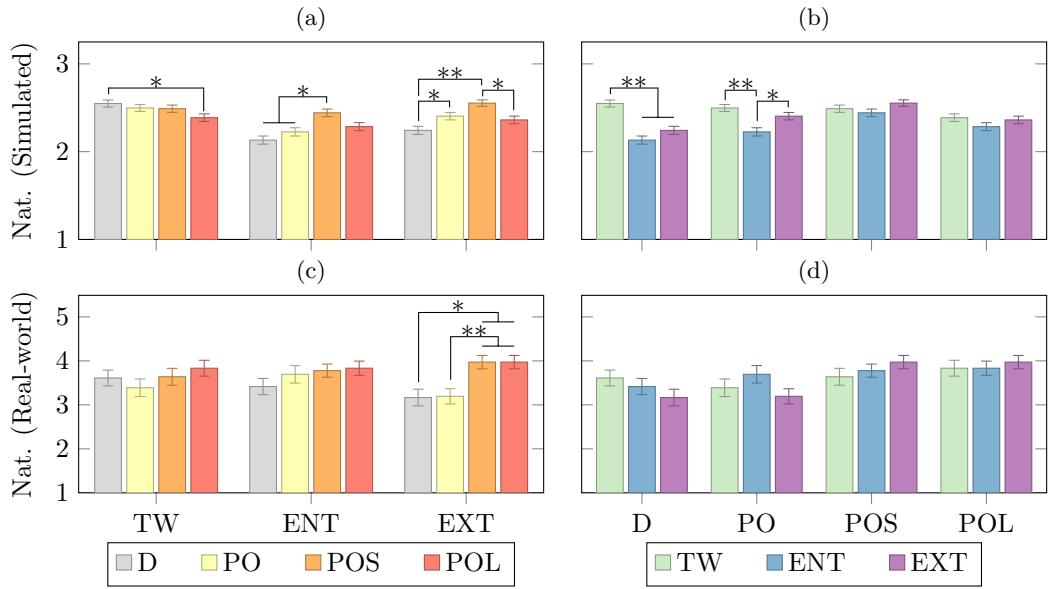

\Cref{fig:cp,fig:pmu,fig:pbi,fig:pred,fig:safe,fig:nat} show the results from the survey questions for each outcome variable. Two-way repeated measures ANOVA tests were first conducted between gaze behaviors and navigation scenarios to confirm whether an interaction effect exists between the two variables. For measures with a statistically significant interaction effect, post-hoc analyses were performed to determine differences between gazes (scenarios) for each scenario (gaze) independently. If the interaction effect was not statistically significant but main effects due to gaze (scenario) were, the data was collapsed along the scenario (gaze) variable before performing post-hoc analyses. Post-hoc analyses were performed using paired sample $t$-tests with Bonferroni corrections. Parametric tests were used as they have been shown to provide similar results to non-parametric results for Likert items~\cite{winter2010}. Post-hoc analysis results are presented in \Cref{fig:cp,fig:pmu,fig:pbi,fig:pred,fig:safe,fig:nat}.

\subsection{Co-Presence}

\subsubsection{Simulated studies}
There was a significant interaction effect between the gaze behavior and the navigation scenario for CP ($F(6,1404)=3.391$, $p<0.05$), therefore, post hoc paired sample $t$-tests were performed for each gaze-scenario pair. 

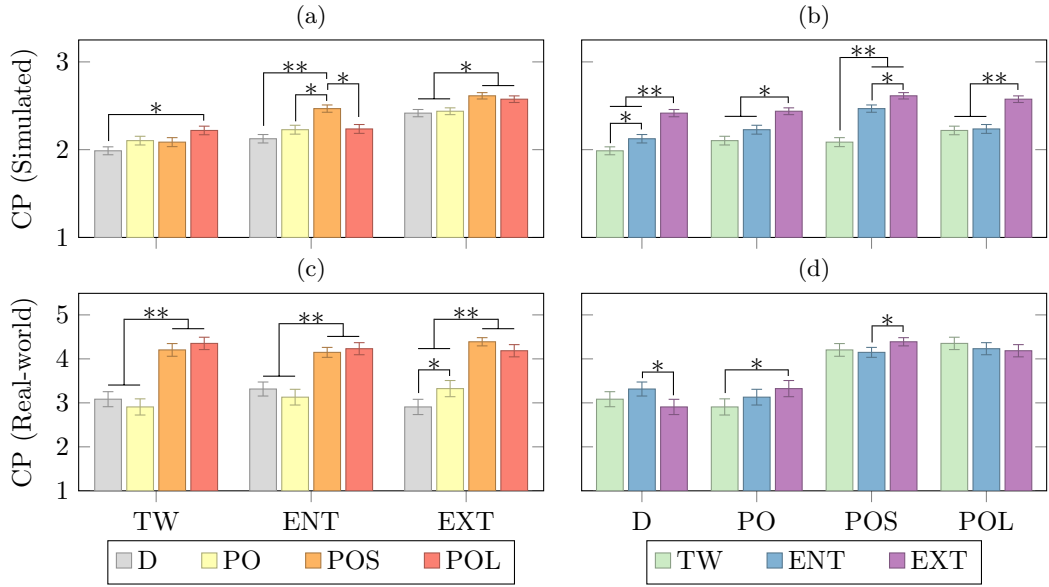
\begin{figure*}[h]
\centering
\begin{tikzpicture}

    \definecolor{a}{RGB}{217,217,217}
    \definecolor{b}{RGB}{255,255,179}
    \definecolor{c}{RGB}{253,180,98}
    \definecolor{d}{RGB}{251,128,114}

    \pgfplotsset{
      /pgfplots/bar  cycle  list/.style={/pgfplots/cycle  list={%
            {a!67!black,fill=a,mark=none},%
            {b!67!black,fill=b,mark=none},%
            {c!67!black,fill=c,mark=none},%
            {d!67!black,fill=d,mark=none},%
         }
      },
    }

    \begin{axis}[
        name=mainplot,
        ybar,
        enlarge x limits=0.25,
        ylabel={CP (Simulated)},
        symbolic x coords={TW,ENT,EXT},
        xtick=data,
        xticklabels={,,},
        bar width=10pt,
        xtick pos=left,
        width = 0.55\columnwidth,
        height = 0.3\columnwidth,
        ymin = 1, ymax = 3.25,
        title=\small(a),
        title style={yshift=-1.25ex,},
    ]
    
    \addplot+ [
        error bars/.cd,
            y dir=both, y explicit,
    ] coordinates {
        (TW,1.9871)  +- (0, 0.04521)
        (ENT,2.1245) +- (0, 0.04806)
        (EXT,2.4163) +- (0, 0.04130)
    };

    \addplot+ [
        error bars/.cd,
            y dir=both, y explicit,
    ] coordinates {
        (TW,2.1030)  +- (0, 0.04959)
        (ENT,2.2275) +- (0, 0.05059)
        (EXT,2.4378) +- (0, 0.03871)
    };
    
    \addplot+ [
        error bars/.cd,
            y dir=both, y explicit,
    ] coordinates {
        (TW,2.0858)  +- (0, 0.05119)
        (ENT,2.4678) +- (0, 0.04161)
        (EXT,2.6137) +- (0, 0.03622)
    };    
    
    \addplot+ [
        error bars/.cd,
            y dir=both, y explicit,
    ] coordinates {
        (TW,2.2189)  +- (0, 0.04890)
        (ENT,2.2361) +- (0, 0.05042)
        (EXT,2.5751) +- (0, 0.03765)
    };
    
    \draw [arrows={Bar[left,width=20pt]-Bar[right]}] 
        (axis cs:TW,2.4) ++(-18pt, 0) -- node[midway, above, yshift=-6.5pt]{*} ++(36pt,0);
        
    \draw [arrows={Bar[left,width=20pt]-Bar[right]}] 
        (axis cs:ENT,2.625) ++(-6pt, 0) -- node[midway, above, yshift=-6.5pt]{*} ++(12pt,0);
        
    \draw [arrows={Bar[left]-Bar[right,width=25pt]}] 
        (axis cs:ENT,2.75) ++(6pt, 0) -- node[midway, above, yshift=-6.5pt]{*} ++(12pt,0);    
        
    \draw [arrows={Bar[left,width=40pt]-Bar[right]}] 
        (axis cs:ENT,2.875) ++(-18pt, 0) -- node[midway, above, yshift=-6.5pt]{**} ++(24pt,0);
        
    \draw [arrows={Bar[left,width=20pt,length=12pt]-Bar[right,width=10pt,length=12pt]}] 
        (axis cs:EXT,2.875) ++(-18pt, 0) -- node[midway, above, yshift=-6.5pt]{*} ++(36pt,0);
    
    \end{axis}
    \definecolor{a}{RGB}{204,235,197}
    \definecolor{b}{RGB}{128,177,211}
    \definecolor{c}{RGB}{188,128,189}
    \begin{axis}[
        name=secondplot,
        at={(mainplot.north east)},
        xshift=0.5cm,
        anchor=north west,    
        ybar,
        enlarge x limits=0.175,
        symbolic x coords={D,PO,POS,POL},
        xtick=data,
        yticklabels={,,},
        xticklabels={,,},
        bar width=10pt,
        xtick pos=left,
        width = 0.55\columnwidth,
        height = 0.3\columnwidth,
        ymin = 1, ymax = 3.25,
        title=\small(b),
        title style={yshift=-1.25ex,},
    ]
    
    \addplot+ [
        error bars/.cd,
            y dir=both, y explicit,
    ] coordinates {
        (D,1.9871)  +- (0, 0.04521)
        (PO,2.1030)  +- (0, 0.04959)
        (POS,2.0858)  +- (0, 0.05119)
        (POL,2.2189)  +- (0, 0.04890)
    };

    \addplot+ [
        error bars/.cd,
            y dir=both, y explicit,
    ] coordinates {
        (D,2.1245) +- (0, 0.04806)
        (PO,2.2275) +- (0, 0.05059)
        (POS,2.4678) +- (0, 0.04161)
        (POL,2.2361) +- (0, 0.05042)
    };
    
    \addplot+ [
        error bars/.cd,
            y dir=both, y explicit,
    ] coordinates {
        (D,2.4163) +- (0, 0.04130)
        (PO,2.4378) +- (0, 0.03871)
        (POS,2.6137) +- (0, 0.03622)
        (POL,2.5751) +- (0, 0.03765)
    };    
    
    \draw [arrows={Bar[left,width=15pt]-Bar[right]}] 
        (axis cs:D,2.3) ++(-12pt, 0) -- node[midway, above, yshift=-6.5pt]{*} ++(12pt,0);
        
    \draw [arrows={Bar[left,width=7.5pt,length=12pt]-Bar[right]}] 
        (axis cs:D,2.6) ++(-12pt, 0) -- node[midway, above, yshift=-6.5pt,xshift=3pt]{**} ++(24pt,0);
        
    \draw [arrows={Bar[left,width=15pt,length=12pt]-Bar[right]}] 
        (axis cs:PO,2.6) ++(-12pt, 0) -- node[midway, above, yshift=-6.5pt,xshift=3pt]{*} ++(24pt,0);
        
    \draw [arrows={Bar[left,width=12.5pt]-Bar[right]}] 
        (axis cs:POS,2.75) ++(0pt, 0) -- node[midway, above, yshift=-6.5pt]{*} ++(12pt,0);

    \draw [arrows={Bar[left,width=55pt]-Bar[right,width=7.5pt,length=12pt]}] 
        (axis cs:POS,3.05) ++(-12pt, 0) -- node[midway, above, yshift=-6.5pt,xshift=-3pt]{**} ++(24pt,0);

    \draw [arrows={Bar[left,width=25pt,length=12pt]-Bar[right]}] 
        (axis cs:POL,2.75) ++(-12pt, 0) -- node[midway, above, yshift=-6.5pt,xshift=3pt]{**} ++(24pt,0);
        
    \end{axis}
    \definecolor{a}{RGB}{217,217,217}
    \definecolor{b}{RGB}{255,255,179}
    \definecolor{c}{RGB}{253,180,98}
    \definecolor{d}{RGB}{251,128,114}
    \begin{axis}[
        at={(mainplot.below south west)},
        yshift=-0.35cm,
        anchor=north west,
        ybar,
        enlarge x limits=0.25,
        legend style={at={(0.5,-0.25)},
        anchor=north,legend columns=-1,
        /tikz/every even column/.append style={column sep=0.5cm}},
        ylabel={CP (Real-world)},
        symbolic x coords={TW,ENT,EXT},
        xtick=data,
        bar width=10pt,
        xtick pos=left,
        width = 0.55\columnwidth,
        height = 0.3\columnwidth,
        ymin = 1, ymax = 5.49,
        legend image code/.code={\draw [#1] (0cm,-0.075cm) rectangle (0.2cm,0.125cm); },
        title=\small(c),
        title style={yshift=-1.25ex,},
    ]
    
    \addplot+ [
        error bars/.cd,
            y dir=both, y explicit,
    ] coordinates {
        (TW,3.0833)  +- (0, 0.1708)
        (ENT,3.3148) +- (0, 0.1593)
        (EXT,2.9074) +- (0, 0.1735)
    };

    \addplot+ [
        error bars/.cd,
            y dir=both, y explicit,
    ] coordinates {
        (TW,2.9074)  +- (0, 0.1838)
        (ENT,3.1296) +- (0, 0.1788)
        (EXT,3.3241) +- (0, 0.1838)
    };
    
    \addplot+ [
        error bars/.cd,
            y dir=both, y explicit,
    ] coordinates {
        (TW,4.2037)  +- (0, 0.1444)
        (ENT,4.1481) +- (0, 0.1130)
        (EXT,4.3889) +- (0, 0.0934)
    };    
    
    \addplot+ [
        error bars/.cd,
            y dir=both, y explicit,
    ] coordinates {
        (TW,4.3519)  +- (0, 0.1405)
        (ENT,4.2315) +- (0, 0.1373)
        (EXT,4.1852) +- (0, 0.1389)
    };
    
    \draw [arrows={Bar[left,width=50pt,length=12pt]-Bar[right,width=7.5pt,length=12pt]}] 
        (axis cs:TW,4.9) ++(-18pt, 0) -- node[midway, above, yshift=-6.5pt]{**} ++(36pt,0);
        
    \draw [arrows={Bar[left,width=40pt,length=12pt]-Bar[right,width=10pt,length=12pt]}] 
        (axis cs:ENT,4.8) ++(-18pt, 0) -- node[midway, above, yshift=-6.5pt]{**} ++(36pt,0);        
        
    \draw [arrows={Bar[left,width=17.5pt]-Bar[right]}] 
        (axis cs:EXT,3.75) ++(-18pt, 0) -- node[midway, above, yshift=-6.5pt]{*} ++(12pt,0);        
        
    \draw [arrows={Bar[left,width=22.5pt,length=12pt]-Bar[right,width=7.5pt,length=12pt]}] 
        (axis cs:EXT,4.9) ++(-18pt, 0) -- node[midway, above, yshift=-6.5pt]{**} ++(36pt,0);                

    \legend{D,PO,POS,POL}
    
    \end{axis}
    \definecolor{a}{RGB}{204,235,197}
    \definecolor{b}{RGB}{128,177,211}
    \definecolor{c}{RGB}{188,128,189}
    \begin{axis}[
        at={(secondplot.below south west)},
        yshift=-0.35cm,
        anchor=north west,  
        ybar,
        enlarge x limits=0.175,
        legend style={at={(0.5,-0.25)},
        anchor=north,legend columns=-1,
        /tikz/every even column/.append style={column sep=0.5cm}},
        symbolic x coords={D,PO,POS,POL},
        xtick=data,
        yticklabels={,,},
        bar width=10pt,
        xtick pos=left,
        width = 0.55\columnwidth,
        height = 0.3\columnwidth,
        ymin = 1, ymax = 5.49,
        legend image code/.code={\draw [#1] (0cm,-0.075cm) rectangle (0.2cm,0.125cm); },
        title=\small(d),
        title style={yshift=-1.25ex,},
    ]
    
    \addplot+ [
        error bars/.cd,
            y dir=both, y explicit,
    ] coordinates {
        (D,3.0833)  +- (0, 0.1708)
        (PO,2.9074)  +- (0, 0.1838)
        (POS,4.2037)  +- (0, 0.1444)
        (POL,4.3519)  +- (0, 0.1405)
    };

    \addplot+ [
        error bars/.cd,
            y dir=both, y explicit,
    ] coordinates {
        (D,3.3148) +- (0, 0.1593)
        (PO,3.1296) +- (0, 0.1788)
        (POS,4.1481) +- (0, 0.1130)
        (POL,4.2315) +- (0, 0.1373)
    };
    
    \addplot+ [
        error bars/.cd,
            y dir=both, y explicit,
    ] coordinates {
        (D,2.9074) +- (0, 0.1735)
        (PO,3.3241) +- (0, 0.1838)
        (POS,4.3889) +- (0, 0.0934)
        (POL,4.1852) +- (0, 0.1389)
    };
    
    \legend{TW,ENT,EXT}
    
    \draw [arrows={Bar[left]-Bar[right,width=17.5pt]}] 
        (axis cs:D,3.7) ++(0pt, 0) -- node[midway, above, yshift=-6.5pt]{*} ++(12pt,0);
        
    \draw [arrows={Bar[left,width=17.5pt]-Bar[right]}] 
        (axis cs:PO,3.75) ++(-12pt, 0) -- node[midway, above, yshift=-6.5pt]{*} ++(24pt,0);

    \draw [arrows={Bar[left,width=12.5pt]-Bar[right]}] 
        (axis cs:POS,4.75) ++(0pt, 0) -- node[midway, above, yshift=-6.5pt]{*} ++(12pt,0);
        
    \end{axis}        
\end{tikzpicture}
\caption{Results obtained for Co-Presence (CP) for simulated (top) and real-world (bottom) user studies. Results are shown grouped by gaze behavior (left) and navigation scenario (right). Significant paired differences are indicated with ($*$) for $p<0.05$, and with ($**$) for $p<0.001$.}
\label{fig:cp}
\end{figure*}

Comparing gaze behaviors within each navigation scenario (Fig.~\ref{fig:cp}a), we see that in general, either one or both of the person-orientated gaze behaviors (POS and POL) yield significantly higher CP measures. POL was most favoured for the TW scenario, POS was most favoured for the ENT scenario, and both POL and POS were equally most favoured for the EXT scenario. This supports Hypothesis 1.

Comparing navigation scenarios given each gaze behavior (Fig.~\ref{fig:cp}b), we observe a common trend across all gaze behaviors. In general, people feel higher co-presences in the order of EXT, ENT, then TW, although the difference is not always significant between TW and ENT scenarios -- specificially for PO and POL gazes.

\subsubsection{Real-world studies}

Significant interaction effects were found between gaze behavior and navigation scenario ($F(6,210)=3.599$, $p<0.05$), therefore post hoc paired sample $t$-tests were performed for each gaze-scenario pair. 

Comparing gaze behaviors within each navigation scenario (Fig.~\ref{fig:cp}c), a common observation between all navigation scenarios is that POS and POL gaze behaviors both have significantly better CP compared to both D and PO gaze behaviors. No significant difference was found between POS and POL, nor between D and PO gazes except for the EXT scenario. This is a somewhat similar but more consistent observation compared to the simulated results, hence providing further support for Hypothesis 1.

Comparing navigation scenarios given each gaze behavior (Fig.~\ref{fig:cp}d), user responses to each gaze behavior appear to be influenced differently by the navigation scenarios. Notably, while the EXT scenario has the highest CP when performing the PO or POS gazes, it is the scenario with the lowest CP for the D gaze.

\subsection{Perceived Message Understanding}
\subsubsection{Simulated studies}
There was significant interaction between the gaze behavior and the navigation scenario for PMU metric ($F(6,1404)=4.771$, $p<0.05$), therefore, post hoc paired sample $t$-test was performed for each gaze-scenario pair. 

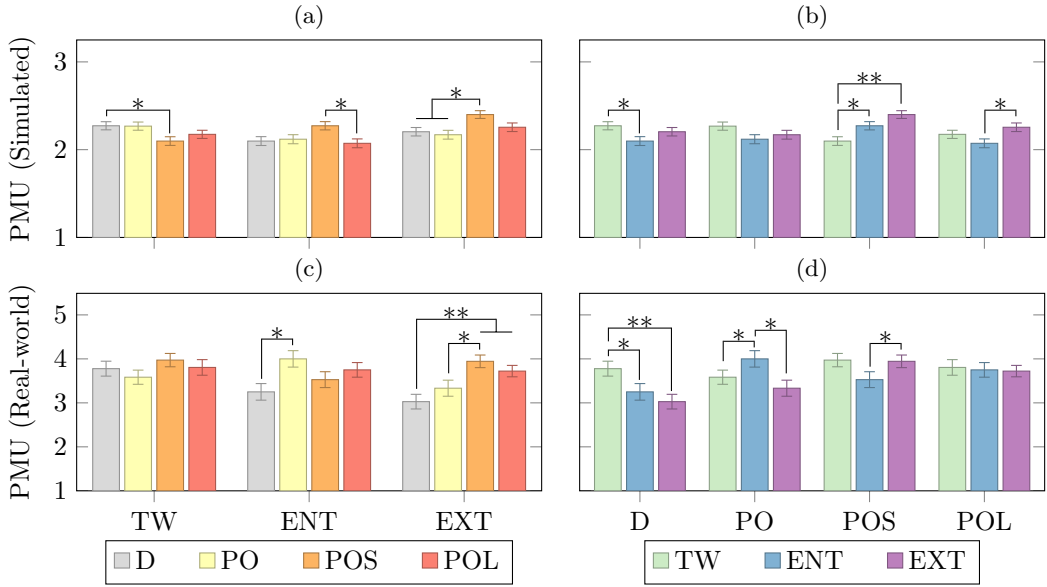
\begin{figure*}[t]
\centering
\begin{tikzpicture}

    \definecolor{a}{RGB}{217,217,217}
    \definecolor{b}{RGB}{255,255,179}
    \definecolor{c}{RGB}{253,180,98}
    \definecolor{d}{RGB}{251,128,114}

    \pgfplotsset{
      /pgfplots/bar  cycle  list/.style={/pgfplots/cycle  list={%
            {a!67!black,fill=a,mark=none},%
            {b!67!black,fill=b,mark=none},%
            {c!67!black,fill=c,mark=none},%
            {d!67!black,fill=d,mark=none},%
         }
      },
    }

    \begin{axis}[
        name=mainplot,
        ybar,
        enlarge x limits=0.25,
        ylabel={PMU (Simulated)},
        symbolic x coords={TW,ENT,EXT},
        xtick=data,
        xticklabels={,,},
        bar width=10pt,
        xtick pos=left,
        width = 0.55\columnwidth,
        height = 0.3\columnwidth,
        ymin = 1, ymax = 3.25,
        title=\small(a),
        title style={yshift=-1.25ex,},
    ]
    
    \addplot+ [
        error bars/.cd,
            y dir=both, y explicit,
    ] coordinates {
        (TW,2.2723)  +- (0, 0.04634)
        (ENT,2.0979) +- (0, 0.05048)
        (EXT,2.2043) +- (0, 0.04818)
    };

    \addplot+ [
        error bars/.cd,
            y dir=both, y explicit,
    ] coordinates {
        (TW,2.2681)  +- (0, 0.04664)
        (ENT,2.1191) +- (0, 0.05153)
        (EXT,2.1702) +- (0, 0.05020)
    };
    
    \addplot+ [
        error bars/.cd,
            y dir=both, y explicit,
    ] coordinates {
        (TW,2.0979)  +- (0, 0.04940)
        (ENT,2.2723) +- (0, 0.04711)
        (EXT,2.4000) +- (0, 0.04430)
    };    
    
    \addplot+ [
        error bars/.cd,
            y dir=both, y explicit,
    ] coordinates {
        (TW,2.1745)  +- (0, 0.04659)
        (ENT,2.0723) +- (0, 0.05031)
        (EXT,2.2553) +- (0, 0.04902)
    };
    
    \draw [arrows={Bar[left]-Bar[right,width=17.5pt]}] 
        (axis cs:TW,2.45) ++(-18pt, 0) -- node[midway, above, yshift=-6.5pt]{*} ++(24pt,0);
        
    \draw [arrows={Bar[left]-Bar[right,width=17.5pt]}] 
        (axis cs:ENT,2.45) ++(6pt, 0) -- node[midway, above, yshift=-6.5pt]{*} ++(12pt,0);
    
    \draw [arrows={Bar[left,width=15pt,length=12pt]-Bar[right]}] 
        (axis cs:EXT,2.575) ++(-18pt, 0) -- node[midway, above, yshift=-6.5pt,xshift=3pt]{*} ++(24pt,0);
    
    \end{axis}
    \definecolor{a}{RGB}{204,235,197}
    \definecolor{b}{RGB}{128,177,211}
    \definecolor{c}{RGB}{188,128,189}        
    \begin{axis}[
        name=secondplot,
        at={(mainplot.north east)},
        xshift=0.5cm,
        anchor=north west,    
        ybar,
        enlarge x limits=0.175,
        symbolic x coords={D,PO,POS,POL},
        xtick=data,
        yticklabels={,,},
        xticklabels={,,},
        bar width=10pt,
        xtick pos=left,
        width = 0.55\columnwidth,
        height = 0.3\columnwidth,
        ymin = 1, ymax = 3.25,
        title=\small(b),
        title style={yshift=-1.25ex,},        
    ]
    
    \addplot+ [
        error bars/.cd,
            y dir=both, y explicit,
    ] coordinates {
        (D,2.2723)  +- (0, 0.04634)
        (PO,2.2681)  +- (0, 0.04664)
        (POS,2.0979)  +- (0, 0.04940)
        (POL,2.1745)  +- (0, 0.04659)
    };

    \addplot+ [
        error bars/.cd,
            y dir=both, y explicit,
    ] coordinates {
        (D,2.0979) +- (0, 0.05048)
        (PO,2.1191) +- (0, 0.05153)
        (POS,2.2723) +- (0, 0.04711)
        (POL,2.0723) +- (0, 0.05031)
    };
    
    \addplot+ [
        error bars/.cd,
            y dir=both, y explicit,
    ] coordinates {
        (D,2.2043) +- (0, 0.04818)
        (PO,2.1702) +- (0, 0.05020)
        (POS,2.4000) +- (0, 0.04430)
        (POL,2.2553) +- (0, 0.04902)
    };    
    
    \draw [arrows={Bar[left]-Bar[right,width=17.5pt]}] 
        (axis cs:D,2.45) ++(-12pt, 0) -- node[midway, above, yshift=-6.5pt]{*} ++(12pt,0);
        
    \draw [arrows={Bar[left,width=15pt]-Bar[right]}] 
        (axis cs:POS,2.45) ++(-12pt, 0) -- node[midway, above, yshift=-6.5pt]{*} ++(12pt,0);
        
    \draw [arrows={Bar[left,width=15pt]-Bar[right,width=15pt]}] 
        (axis cs:POS,2.75) ++(-12pt, 0) -- node[midway, above, yshift=-6.5pt]{**} ++(24pt,0);
        
    \draw [arrows={Bar[left,width=17.5pt]-Bar[right]}] 
        (axis cs:POL,2.45) ++(0pt, 0) -- node[midway, above, yshift=-6.5pt]{*} ++(12pt,0);
        
    \end{axis}
    \definecolor{a}{RGB}{217,217,217}
    \definecolor{b}{RGB}{255,255,179}
    \definecolor{c}{RGB}{253,180,98}
    \definecolor{d}{RGB}{251,128,114}    
    \begin{axis}[
        at={(mainplot.below south west)},
        yshift=-0.35cm,
        anchor=north west,
        ybar,
        enlarge x limits=0.25,
        legend style={at={(0.5,-0.25)},
        anchor=north,legend columns=-1,
        /tikz/every even column/.append style={column sep=0.5cm}},
        ylabel={PMU (Real-world)},
        symbolic x coords={TW,ENT,EXT},
        xtick=data,
        bar width=10pt,
        xtick pos=left,
        width = 0.55\columnwidth,
        height = 0.3\columnwidth,
        ymin = 1, ymax = 5.49,
        legend image code/.code={\draw [#1] (0cm,-0.075cm) rectangle (0.2cm,0.125cm); },
        title=\small(c),
        title style={yshift=-1.25ex,},
    ]
    
    \addplot+ [
        error bars/.cd,
            y dir=both, y explicit,
    ] coordinates {
        (TW,3.7778)  +- (0, 0.1696)
        (ENT,3.2500) +- (0, 0.1885)
        (EXT,3.0278) +- (0, 0.1666)
    };

    \addplot+ [
        error bars/.cd,
            y dir=both, y explicit,
    ] coordinates {
        (TW,3.5833)  +- (0, 0.1612)
        (ENT,4.0000) +- (0, 0.1869)
        (EXT,3.3333) +- (0, 0.1826)
    };
    
    \addplot+ [
        error bars/.cd,
            y dir=both, y explicit,
    ] coordinates {
        (TW,3.9722)  +- (0, 0.1516)
        (ENT,3.5278) +- (0, 0.1803)
        (EXT,3.9444) +- (0, 0.1433)
    };    
    
    \addplot+ [
        error bars/.cd,
            y dir=both, y explicit,
    ] coordinates {
        (TW,3.8056)  +- (0, 0.1774)
        (ENT,3.7500) +- (0, 0.1661)
        (EXT,3.7222) +- (0, 0.1298)
    };
    
    \draw [arrows={Bar[left,width=30pt]-Bar[right]}] 
        (axis cs:ENT,4.45) ++(-18pt, 0) -- node[midway, above, yshift=-6.5pt]{*} ++(12pt,0);
        
    \draw [arrows={Bar[left,width=22.5pt]-Bar[right]}] 
        (axis cs:EXT,4.35) ++(-6pt, 0) -- node[midway, above, yshift=-6.5pt]{*} ++(12pt,0);        
        
    \draw [arrows={Bar[left,width=52.5pt]-Bar[right,width=10pt,length=12pt]}] 
        (axis cs:EXT,4.9) ++(-18pt, 0) -- node[midway, above, yshift=-6.5pt,xshift=-3pt]{**} ++(36pt,0);                 

    \legend{D,PO,POS,POL}
    
    \end{axis}
    \definecolor{a}{RGB}{204,235,197}
    \definecolor{b}{RGB}{128,177,211}
    \definecolor{c}{RGB}{188,128,189}        
    \begin{axis}[
        at={(secondplot.below south west)},
        yshift=-0.35cm,
        anchor=north west,  
        ybar,
        enlarge x limits=0.175,
        legend style={at={(0.5,-0.25)},
        anchor=north,legend columns=-1,
        /tikz/every even column/.append style={column sep=0.5cm}},
        symbolic x coords={D,PO,POS,POL},
        xtick=data,
        yticklabels={,,},
        bar width=10pt,
        xtick pos=left,
        width = 0.55\columnwidth,
        height = 0.3\columnwidth,
        ymin = 1, ymax = 5.49,
        legend image code/.code={\draw [#1] (0cm,-0.075cm) rectangle (0.2cm,0.125cm); },
        title=\small(d),
        title style={yshift=-1.25ex,},
    ]
    
    \addplot+ [
        error bars/.cd,
            y dir=both, y explicit,
    ] coordinates {
        (D,3.7778)  +- (0, 0.1696)
        (PO,3.5833)  +- (0, 0.1612)
        (POS,3.9722)  +- (0, 0.1516)
        (POL,3.8056)  +- (0, 0.1774)
    };

    \addplot+ [
        error bars/.cd,
            y dir=both, y explicit,
    ] coordinates {
        (D,3.2500) +- (0, 0.1885)
        (PO,4.0000) +- (0, 0.1869)
        (POS,3.5278) +- (0, 0.1803)
        (POL,3.7500) +- (0, 0.1661)
    };
    
    \addplot+ [
        error bars/.cd,
            y dir=both, y explicit,
    ] coordinates {
        (D,3.0278) +- (0, 0.1666)
        (PO,3.3333) +- (0, 0.1826)
        (POS,3.9444) +- (0, 0.1433)
        (POL,3.7222) +- (0, 0.1298)
    };
    
    \legend{TW,ENT,EXT}
    
    \draw [arrows={Bar[left]-Bar[right,width=22.5pt]}] 
        (axis cs:D,4.2) ++(-12pt, 0) -- node[midway, above, yshift=-6.5pt]{*} ++(12pt,0);
        
    \draw [arrows={Bar[left,width=10pt]-Bar[right,width=47.5pt]}] 
        (axis cs:D,4.7) ++(-12pt, 0) -- node[midway, above, yshift=-6.5pt]{**} ++(24pt,0);    

    \draw [arrows={Bar[left,width=15pt]-Bar[right]}] 
        (axis cs:PO,4.4) ++(-12pt, 0) -- node[midway, above, yshift=-6.5pt]{*} ++(12pt,0);
        
    \draw [arrows={Bar[left]-Bar[right,width=32.5pt]}] 
        (axis cs:PO,4.65) ++(0pt, 0) -- node[midway, above, yshift=-6.5pt]{*} ++(12pt,0);    

    \draw [arrows={Bar[left,width=17.5pt]-Bar[right]}] 
        (axis cs:POS,4.35) ++(0pt, 0) -- node[midway, above, yshift=-6.5pt]{*} ++(12pt,0);

    \end{axis}        
\end{tikzpicture}
\caption{Results obtained for Perceived Message Understanding (PMU) for simulated (top) and real-world (bottom) user studies. Results are shown grouped by gaze behavior (left) and navigation scenario (right). Significant paired differences are indicated with ($*$) for $p<0.05$, and with ($**$) for $p<0.001$.}
\label{fig:pmu}
\end{figure*}

Comparing gaze behaviors within each navigation scenario (Fig.~\ref{fig:pmu}a), we observe that PMU is significantly higher for the D gaze compared to the POS gaze for the TW scenario, which is in opposition to Hypothesis 1. However, for the ENT and EXT scenarios, the POS gaze is rated to have the best PMU, which is in support for Hypothesis 1.

Comparing navigation scenarios given each gaze behavior (Fig.~\ref{fig:pmu}b), a couple of observations can be made. Among D gazes, the TW scenario yielded a significantly better PMU compared to the ENT scenario. For POS gazes however, the TW scenario yielded significantly worse PMU compared to the other two scenarios. For POL gazes, the EXT scenario yielded significantly better PMU compared to the ENT scenario.

\subsubsection{Real-world studies}
Significant interaction effects were found between gaze behavior and navigation scenario ($F(6,210)=5.370$, $p<0.05$), therefore post hoc paired sample $t$-tests were performed for each gaze-scenario pair. 

Comparing gaze behaviors within each navigation scenario (Fig.~\ref{fig:pmu}c), for the ENT scenario, PO is rated to have the best PMU, and is significantly better compared to the D gaze. For the EXT scenario, the D gaze also has the worst PMU, however the person-oriented gaze behaviors (POS and POL) are perceived to have the best PMU. Hypothesis 1 is therefore only supported for the EXT scenario.

Comparing navigation scenarios given each gaze behavior (Fig.~\ref{fig:pmu}d), we see that for the D gaze, the TW scenario had the best PMU compared to the other two scenarios. For the PO gaze, the robot had the best PMU for the ENT scenario compared to the other two scenarios. For the POS gaze, the EXT scenario was found to have significantly better PMU compared to the ENT scenario. 

\subsection{Perceived Behavioral Interdependence}
\subsubsection{Simulated studies}
There was significant interaction between the gaze behavior and the navigation scenario for the PBI metric ($F(6,1404)=3.354$, $p<0.05$), therefore, post hoc paired sample $t$-test was performed for each gaze-scenario pair. 

\begin{figure*}[t]
\centering
\begin{tikzpicture}

    \definecolor{a}{RGB}{217,217,217}
    \definecolor{b}{RGB}{255,255,179}
    \definecolor{c}{RGB}{253,180,98}
    \definecolor{d}{RGB}{251,128,114}

    \pgfplotsset{
      /pgfplots/bar  cycle  list/.style={/pgfplots/cycle  list={%
            {a!67!black,fill=a,mark=none},%
            {b!67!black,fill=b,mark=none},%
            {c!67!black,fill=c,mark=none},%
            {d!67!black,fill=d,mark=none},%
         }
      },
    }

    \begin{axis}[
        name=mainplot,
        ybar,
        enlarge x limits=0.25,
        ylabel={PBI (Simulated)},
        symbolic x coords={TW,ENT,EXT},
        xtick=data,
        xticklabels={,,},
        bar width=10pt,
        xtick pos=left,
        width = 0.55\columnwidth,
        height = 0.3\columnwidth,
        ymin = 1, ymax = 3.25,
        title=\small(a),
        title style={yshift=-1.25ex,},        
    ]
    
    \addplot+ [
        error bars/.cd,
            y dir=both, y explicit,
    ] coordinates {
        (TW,2.0128)  +- (0, 0.05016)
        (ENT,2.0170) +- (0, 0.04775)
        (EXT,2.0936) +- (0, 0.04888)
    };

    \addplot+ [
        error bars/.cd,
            y dir=both, y explicit,
    ] coordinates {
        (TW,1.9702)  +- (0, 0.04904)
        (ENT,1.9830) +- (0, 0.04888)
        (EXT,2.2340) +- (0, 0.04129)
    };
    
    \addplot+ [
        error bars/.cd,
            y dir=both, y explicit,
    ] coordinates {
        (TW,2.0426)  +- (0, 0.05027)
        (ENT,2.2213) +- (0, 0.04972)
        (EXT,2.4298) +- (0, 0.03983)
    };    
    
    \addplot+ [
        error bars/.cd,
            y dir=both, y explicit,
    ] coordinates {
        (TW,2.0426)  +- (0, 0.04769)
        (ENT,2.1957) +- (0, 0.04870)
        (EXT,2.4638) +- (0, 0.03956)
    };
    
    \draw [arrows={Bar[left,width=25pt,length=12pt]-Bar[right,width=10pt,length=12pt]}] 
        (axis cs:ENT,2.6) ++(-18pt, 0) -- node[midway, above, yshift=-6.5pt]{*} ++(36pt,0);    
        
    \draw [arrows={Bar[left,width=15pt]-Bar[right]}] 
        (axis cs:EXT,2.575) ++(-6pt, 0) -- node[midway, above, yshift=-6.5pt]{*} ++(12pt,0);
        
    \draw [arrows={Bar[left,width=10pt]-Bar[right,width=15pt]}] 
        (axis cs:EXT,2.775) ++(-6pt, 0) -- node[midway, above, yshift=-6.5pt]{**} ++(24pt,0);
        
    \draw [arrows={Bar[left,width=55pt]-Bar[right,width=7.5pt,length=12pt]}] 
        (axis cs:EXT,3.05) ++(-18pt, 0) -- node[midway, above, yshift=-6.5pt, xshift=-3pt]{**} ++(36pt,0);
    
    \end{axis}
    \definecolor{a}{RGB}{204,235,197}
    \definecolor{b}{RGB}{128,177,211}
    \definecolor{c}{RGB}{188,128,189}    
    \begin{axis}[
        name=secondplot,
        at={(mainplot.north east)},
        xshift=0.5cm,
        anchor=north west,    
        ybar,
        enlarge x limits=0.175,
        symbolic x coords={D,PO,POS,POL},
        xtick=data,
        xticklabels={,,},
        yticklabels={,,},
        bar width=10pt,
        xtick pos=left,
        width = 0.55\columnwidth,
        height = 0.3\columnwidth,
        ymin = 1, ymax = 3.25,
        title=\small(b),
        title style={yshift=-1.25ex,},        
    ]
    
    \addplot+ [
        error bars/.cd,
            y dir=both, y explicit,
    ] coordinates {
        (D,2.0128)  +- (0, 0.05016)
        (PO,1.9702)  +- (0, 0.04904)
        (POS,2.0426)  +- (0, 0.05027)
        (POL,2.0426)  +- (0, 0.04769)
    };

    \addplot+ [
        error bars/.cd,
            y dir=both, y explicit,
    ] coordinates {
        (D,2.0170) +- (0, 0.04775)
        (PO,1.9830) +- (0, 0.04888)
        (POS,2.2213) +- (0, 0.04972)
        (POL,2.1957) +- (0, 0.04870)
    };
    
    \addplot+ [
        error bars/.cd,
            y dir=both, y explicit,
    ] coordinates {
        (D,2.0936) +- (0, 0.04888)
        (PO,2.2340) +- (0, 0.04129)
        (POS,2.4298) +- (0, 0.03983)
        (POL,2.4638) +- (0, 0.03956)
    };
    
    \draw [arrows={Bar[left,width=20pt,length=12pt]-Bar[right]}] 
        (axis cs:PO,2.4) ++(-12pt, 0) -- node[midway, above, yshift=-6.5pt, xshift=3pt]{*} ++(24pt,0);
        
    \draw [arrows={Bar[left,width=17.5pt]-Bar[right]}] 
        (axis cs:POS,2.4) ++(-12pt, 0) -- node[midway, above, yshift=-6.5pt]{*} ++(12pt,0);
        
    \draw [arrows={Bar[left,width=10pt]-Bar[right]}] 
        (axis cs:POS,2.6) ++(0pt, 0) -- node[midway, above, yshift=-6.5pt]{*} ++(12pt,0);
        
    \draw [arrows={Bar[left,width=25pt]-Bar[right,width=12.5pt]}] 
        (axis cs:POS,2.85) ++(-12pt, 0) -- node[midway, above, yshift=-6.5pt]{**} ++(24pt,0);
        
    \draw [arrows={Bar[left,width=20pt,length=12pt]-Bar[right]}] 
        (axis cs:POL,2.65) ++(-12pt, 0) -- node[midway, above, yshift=-6.5pt, xshift=3pt]{*} ++(24pt,0);
        
    \end{axis}
    \definecolor{a}{RGB}{217,217,217}
    \definecolor{b}{RGB}{255,255,179}
    \definecolor{c}{RGB}{253,180,98}
    \definecolor{d}{RGB}{251,128,114}    
    \begin{axis}[
        at={(mainplot.below south west)},
        yshift=-0.35cm,
        anchor=north west,
        ybar,
        enlarge x limits=0.25,
        legend style={at={(0.5,-0.25)},
        anchor=north,legend columns=-1,
        /tikz/every even column/.append style={column sep=0.5cm}},
        ylabel={PBI (Real-world)},
        symbolic x coords={TW,ENT,EXT},
        xtick=data,
        bar width=10pt,
        xtick pos=left,
        width = 0.55\columnwidth,
        height = 0.3\columnwidth,
        ymin = 1, ymax = 5.49,
        legend image code/.code={\draw [#1] (0cm,-0.075cm) rectangle (0.2cm,0.125cm); },
        title=\small(c),
        title style={yshift=-1.25ex,},
    ]
    
    \addplot+ [
        error bars/.cd,
            y dir=both, y explicit,
    ] coordinates {
        (TW,2.8611)  +- (0, 0.1875)
        (ENT,2.6944) +- (0, 0.1585)
        (EXT,2.5833) +- (0, 0.1967)
    };

    \addplot+ [
        error bars/.cd,
            y dir=both, y explicit,
    ] coordinates {
        (TW,2.6944)  +- (0, 0.1861)
        (ENT,2.9167) +- (0, 0.1842)
        (EXT,2.8889) +- (0, 0.1901)
    };
    
    \addplot+ [
        error bars/.cd,
            y dir=both, y explicit,
    ] coordinates {
        (TW,3.5833)  +- (0, 0.1562)
        (ENT,3.3611) +- (0, 0.1499)
        (EXT,3.8056) +- (0, 0.1481)
    };    
    
    \addplot+ [
        error bars/.cd,
            y dir=both, y explicit,
    ] coordinates {
        (TW,3.5833)  +- (0, 0.1661)
        (ENT,3.4722) +- (0, 0.1568)
        (EXT,3.6111) +- (0, 0.1557)
    };

    \draw [arrows={Bar[left,width=30pt,length=12pt]-Bar[right,width=7.5pt,length=12pt]}] 
        (axis cs:TW,4.25) ++(-18pt, 0) -- node[midway, above, yshift=-6.5pt]{**} ++(36pt,0);
        
    \draw [arrows={Bar[left,width=25pt,length=12pt]-Bar[right,width=7.5pt,length=12pt]}] 
        (axis cs:ENT,4.05) ++(-18pt, 0) -- node[midway, above, yshift=-6.5pt]{**} ++(36pt,0);    
        
    \draw [arrows={Bar[left,width=40pt,length=12pt]-Bar[right,width=7.5pt,length=12pt]}] 
        (axis cs:EXT,4.45) ++(-18pt, 0) -- node[midway, above, yshift=-6.5pt]{**} ++(36pt,0);            
      
    \legend{D,PO,POS,POL}
    
    \end{axis}
    \definecolor{a}{RGB}{204,235,197}
    \definecolor{b}{RGB}{128,177,211}
    \definecolor{c}{RGB}{188,128,189}    
    \begin{axis}[
        at={(secondplot.below south west)},
        yshift=-0.35cm,
        anchor=north west,  
        ybar,
        enlarge x limits=0.175,
        legend style={at={(0.5,-0.25)},
        anchor=north,legend columns=-1,
        /tikz/every even column/.append style={column sep=0.5cm}},
        symbolic x coords={D,PO,POS,POL},
        xtick=data,
        yticklabels={,,},
        bar width=10pt,
        xtick pos=left,
        width = 0.55\columnwidth,
        height = 0.3\columnwidth,
        ymin = 1, ymax = 5.49,
        legend image code/.code={\draw [#1] (0cm,-0.075cm) rectangle (0.2cm,0.125cm); },
        title=\small(d),
        title style={yshift=-1.25ex,},
    ]
    
    \addplot+ [
        error bars/.cd,
            y dir=both, y explicit,
    ] coordinates {
        (D,2.8611)  +- (0, 0.1875)
        (PO,2.6944)  +- (0, 0.1861)
        (POS,3.5833)  +- (0, 0.1562)
        (POL,3.5833)  +- (0, 0.1661)
    };

    \addplot+ [
        error bars/.cd,
            y dir=both, y explicit,
    ] coordinates {
        (D,2.6944) +- (0, 0.1585)
        (PO,2.9167) +- (0, 0.1842)
        (POS,3.3611) +- (0, 0.1499)
        (POL,3.4722) +- (0, 0.1568)
    };
    
    \addplot+ [
        error bars/.cd,
            y dir=both, y explicit,
    ] coordinates {
        (D,2.5833) +- (0, 0.1967)
        (PO,2.8889) +- (0, 0.1901)
        (POS,3.8056) +- (0, 0.1481)
        (POL,3.6111) +- (0, 0.1557)
    };

    \legend{TW,ENT,EXT}

    \end{axis}        
\end{tikzpicture}
\caption{Results obtained for Perceived Behavioral Interdependence (PBI) for simulated (top) and real-world (bottom) user studies. Results are shown grouped by gaze behavior (left) and navigation scenario (right). Significant paired differences are indicated with ($*$) for $p<0.05$, and with ($**$) for $p<0.001$.}
\label{fig:pbi}
\end{figure*}
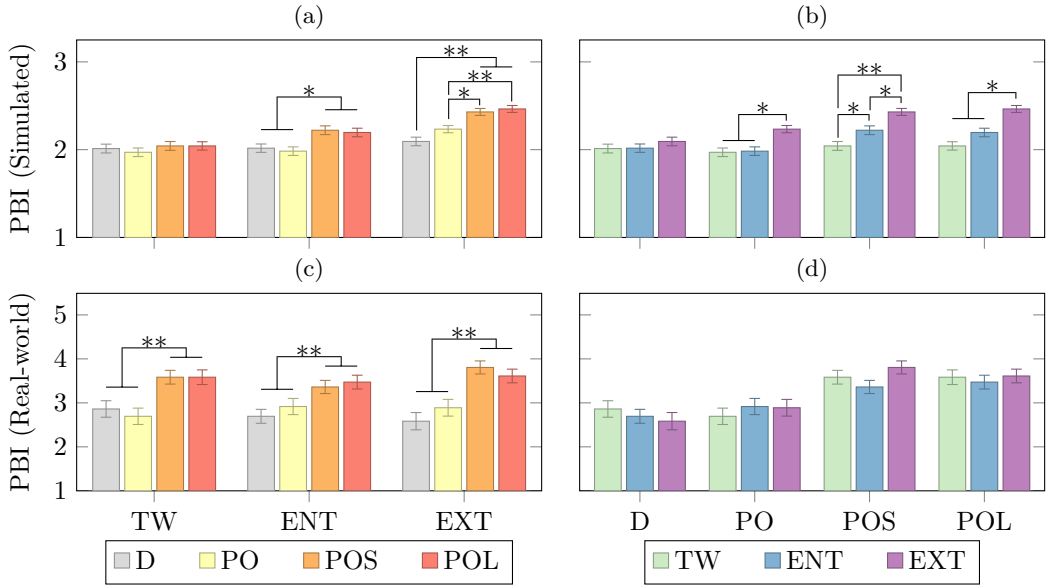

Comparing gaze behaviors within each navigation scenario (Fig.~\ref{fig:pbi}a), we found that PBI with person-oriented gaze behaviors (POS and POL) was significantly higher compared to both D and PO gazes for ENT and EXT scenarios. However, there is no statistically significant difference between any gaze behaviors for the TW scenario. This supports Hypothesis 1 predicting improved social presence metrics when person-oriented gaze behaviors are used, but for the ENT and EXT scenarios only.

Comparing navigation scenarios given each gaze behavior (Fig.~\ref{fig:pbi}b), there is a general trend where PBI is highest in the EXT scenario, followed by the ENT scenario, followed by the TW scenario, when path or person oriented gazes are used (PO, POS, POL). However, this trend is not observed with the D gaze behavior.

\subsubsection{Real-world studies}
There was no significant interaction effect between gaze behavior and navigation scenario ($F(6,210)=1.679$, $p=0.13$). There were significant main effects exhibited by gaze behaviors ($F(3,105)=29.041$, $p<0.05$), but not by navigational scenarios ($F(2,70)=0.616$, $p=0.54$). Therefore post hoc paired sample $t$-tests were performed for investigate the impact of gaze behaviors after collapsing results in the scenario variable.

Comparing between gaze behaviors (Fig.~\ref{fig:pbi}c), a similar observation to simulated study results is made where PBI is significantly higher for person-oriented gaze behaviors (POS and POL) compared to D and PO gazes. However unlike the simulated study results, this holds for all navigation scenarios, fully supporting Hypothesis 1.

\subsection{Predictability}
\subsubsection{Simulated studies}
There was significant interaction between the gaze behavior and the navigation scenario for the Predictability metric ($F(6,1404)=3.247$, $p<0.05$), therefore, post hoc paired sample $t$-test was performed for each gaze-scenario pair. 

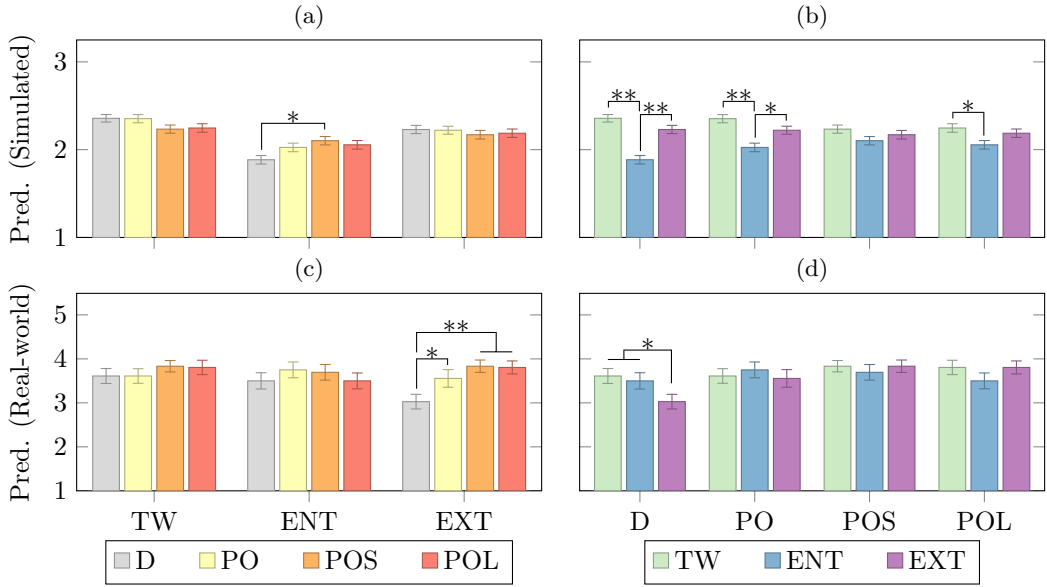
\begin{figure*}[t]
\centering
\begin{tikzpicture}

    \definecolor{a}{RGB}{217,217,217}
    \definecolor{b}{RGB}{255,255,179}
    \definecolor{c}{RGB}{253,180,98}
    \definecolor{d}{RGB}{251,128,114}

    \pgfplotsset{
      /pgfplots/bar  cycle  list/.style={/pgfplots/cycle  list={%
            {a!67!black,fill=a,mark=none},%
            {b!67!black,fill=b,mark=none},%
            {c!67!black,fill=c,mark=none},%
            {d!67!black,fill=d,mark=none},%
         }
      },
    }

    \begin{axis}[
        name=mainplot,
        ybar,
        enlarge x limits=0.25,
        ylabel={Pred. (Simulated)},
        symbolic x coords={TW,ENT,EXT},
        xtick=data,
        xticklabels={,,},
        bar width=10pt,
        xtick pos=left,
        width = 0.55\columnwidth,
        height = 0.3\columnwidth,
        ymin = 1, ymax = 3.25,
        title=\small(a),
        title style={yshift=-1.25ex,},        
    ]
    
    \addplot+ [
        error bars/.cd,
            y dir=both, y explicit,
    ] coordinates {
        (TW,2.3574)  +- (0, 0.04297)
        (ENT,1.8851) +- (0, 0.04850)
        (EXT,2.2298) +- (0, 0.04692)
    };

    \addplot+ [
        error bars/.cd,
            y dir=both, y explicit,
    ] coordinates {
        (TW,2.3532)  +- (0, 0.04655)
        (ENT,2.0255) +- (0, 0.04886)
        (EXT,2.2213) +- (0, 0.04553)
    };
    
    \addplot+ [
        error bars/.cd,
            y dir=both, y explicit,
    ] coordinates {
        (TW,2.2340)  +- (0, 0.04664)
        (ENT,2.1021) +- (0, 0.04768)
        (EXT,2.1702) +- (0, 0.04911)
    };    
    
    \addplot+ [
        error bars/.cd,
            y dir=both, y explicit,
    ] coordinates {
        (TW,2.2468)  +- (0, 0.04846)
        (ENT,2.0553) +- (0, 0.04894)
        (EXT,2.1872) +- (0, 0.04810)
    };
    
    \draw [arrows={Bar[left,width=20pt]-Bar[right]}] 
        (axis cs:ENT,2.3) ++(-18pt, 0) -- node[midway, above, yshift=-6.5pt]{*} ++(24pt,0);    

    \end{axis}
    \definecolor{a}{RGB}{204,235,197}
    \definecolor{b}{RGB}{128,177,211}
    \definecolor{c}{RGB}{188,128,189}        
    \begin{axis}[
        name=secondplot,
        at={(mainplot.north east)},
        xshift=0.5cm,
        anchor=north west,    
        ybar,
        enlarge x limits=0.175,
        symbolic x coords={D,PO,POS,POL},
        xtick=data,
        xticklabels={,,},
        yticklabels={,,},
        bar width=10pt,
        xtick pos=left,
        width = 0.55\columnwidth,
        height = 0.3\columnwidth,
        ymin = 1, ymax = 3.25,
        title=\small(b),
        title style={yshift=-1.25ex,},        
    ]
    
    \addplot+ [
        error bars/.cd,
            y dir=both, y explicit,
    ] coordinates {
        (D,2.3574)  +- (0, 0.04297)
        (PO,2.3532)  +- (0, 0.04655)
        (POS,2.2340)  +- (0, 0.04664)
        (POL,2.2468)  +- (0, 0.04846)
    };

    \addplot+ [
        error bars/.cd,
            y dir=both, y explicit,
    ] coordinates {
        (D,1.8851) +- (0, 0.04850)
        (PO,2.0255) +- (0, 0.04886)
        (POS,2.1021) +- (0, 0.04768)
        (POL,2.0553) +- (0, 0.04894)
    };
    
    \addplot+ [
        error bars/.cd,
            y dir=both, y explicit,
    ] coordinates {
        (D,2.2298) +- (0, 0.04692)
        (PO,2.2213) +- (0, 0.04553)
        (POS,2.1702) +- (0, 0.04911)
        (POL,2.1872) +- (0, 0.04810)
    };

    \draw [arrows={Bar[left]-Bar[right,width=7.5pt]}] 
        (axis cs:D,2.55) ++(-12pt, 0) -- node[midway, above, yshift=-6.5pt]{**} ++(12pt,0);
        
    \draw [arrows={Bar[left,width=27.5pt]-Bar[right]}] 
        (axis cs:D,2.4) ++(0pt, 0) -- node[midway, above, yshift=-6.5pt]{**} ++(12pt,0);
        
    \draw [arrows={Bar[left]-Bar[right,width=7.5pt]}] 
        (axis cs:PO,2.55) ++(-12pt, 0) -- node[midway, above, yshift=-6.5pt]{**} ++(12pt,0);
        
    \draw [arrows={Bar[left,width=17.5pt]-Bar[right]}] 
        (axis cs:PO,2.4) ++(0pt, 0) -- node[midway, above, yshift=-6.5pt]{*} ++(12pt,0);
        
    \draw [arrows={Bar[left]-Bar[right,width=17.5pt]}] 
        (axis cs:POL,2.425) ++(-12pt, 0) -- node[midway, above, yshift=-6.5pt]{*} ++(12pt,0);
        
    \end{axis}
    \definecolor{a}{RGB}{217,217,217}
    \definecolor{b}{RGB}{255,255,179}
    \definecolor{c}{RGB}{253,180,98}
    \definecolor{d}{RGB}{251,128,114}    
    \begin{axis}[
        at={(mainplot.below south west)},
        yshift=-0.35cm,
        anchor=north west,
        ybar,
        enlarge x limits=0.25,
        legend style={at={(0.5,-0.25)},
        anchor=north,legend columns=-1,
        /tikz/every even column/.append style={column sep=0.5cm}},
        ylabel={Pred. (Real-world)},
        symbolic x coords={TW,ENT,EXT},
        xtick=data,
        bar width=10pt,
        xtick pos=left,
        width = 0.55\columnwidth,
        height = 0.3\columnwidth,
        ymin = 1, ymax = 5.49,
        legend image code/.code={\draw [#1] (0cm,-0.075cm) rectangle (0.2cm,0.125cm); },
        title=\small(c),
        title style={yshift=-1.25ex,},        
    ]
    
    \addplot+ [
        error bars/.cd,
            y dir=both, y explicit,
    ] coordinates {
        (TW,3.6111)  +- (0, 0.1703)
        (ENT,3.5000) +- (0, 0.1847)
        (EXT,3.0278) +- (0, 0.1666)
    };
    
    \addplot+ [
        error bars/.cd,
            y dir=both, y explicit,
    ] coordinates {
        (TW,3.6111)  +- (0, 0.1656)
        (ENT,3.7500) +- (0, 0.1798)
        (EXT,3.5556) +- (0, 0.2010)
    };    
    
    \addplot+ [
        error bars/.cd,
            y dir=both, y explicit,
    ] coordinates {
        (TW,3.8333)  +- (0, 0.1291)
        (ENT,3.6944) +- (0, 0.1774)
        (EXT,3.8333) +- (0, 0.1409)
    };
    
    \addplot+ [
        error bars/.cd,
            y dir=both, y explicit,
    ] coordinates {
        (TW,3.8056)  +- (0, 0.1634)
        (ENT,3.5000) +- (0, 0.1804)
        (EXT,3.8056) +- (0, 0.1481)
    };    
      
    \legend{D,PO,POS,POL}
    
    \draw [arrows={Bar[left,width=22.5pt]-Bar[right]}] 
        (axis cs:EXT,4.0) ++(-18pt, 0) -- node[midway, above, yshift=-6.5pt]{*} ++(12pt,0);
        
    \draw [arrows={Bar[left,width=15pt]-Bar[right,width=15pt,length=12pt]}] 
        (axis cs:EXT,4.6) ++(-18pt, 0) -- node[midway, above, yshift=-6.5pt,xshift=-3pt]{**} ++(36pt,0);        
    
    \end{axis}
    \definecolor{a}{RGB}{204,235,197}
    \definecolor{b}{RGB}{128,177,211}
    \definecolor{c}{RGB}{188,128,189}        
    \begin{axis}[
        at={(secondplot.below south west)},
        yshift=-0.35cm,
        anchor=north west,  
        ybar,
        enlarge x limits=0.175,
        legend style={at={(0.5,-0.25)},
        anchor=north,legend columns=-1,
        /tikz/every even column/.append style={column sep=0.5cm}},
        symbolic x coords={D,PO,POS,POL},
        xtick=data,
        yticklabels={,,},
        bar width=10pt,
        xtick pos=left,
        width = 0.55\columnwidth,
        height = 0.3\columnwidth,
        ymin = 1, ymax = 5.49,
        legend image code/.code={\draw [#1] (0cm,-0.075cm) rectangle (0.2cm,0.125cm); },
        title=\small(d),
        title style={yshift=-1.25ex,},        
    ]
    
    \addplot+ [
        error bars/.cd,
            y dir=both, y explicit,
    ] coordinates {
        (D,3.6111)  +- (0, 0.1703)
        (PO,3.6111)  +- (0, 0.1656)
        (POS,3.8333)  +- (0, 0.1291)
        (POL,3.8056)  +- (0, 0.1634)
    };

    \addplot+ [
        error bars/.cd,
            y dir=both, y explicit,
    ] coordinates {
        (D,3.5000) +- (0, 0.1847)
        (PO,3.7500) +- (0, 0.1798)
        (POS,3.6944) +- (0, 0.1774)
        (POL,3.5000) +- (0, 0.1804)
    };
    
    \addplot+ [
        error bars/.cd,
            y dir=both, y explicit,
    ] coordinates {
        (D,3.0278) +- (0, 0.1666)
        (PO,3.5556) +- (0, 0.2010)
        (POS,3.8333) +- (0, 0.1409)
        (POL,3.8056) +- (0, 0.1481)
    };
    
    \legend{TW,ENT,EXT}
    
    \draw [arrows={Bar[left,width=7.5pt,length=12pt]-Bar[right,width=25pt]}] 
        (axis cs:D,4.2) ++(-12pt, 0) -- node[midway, above, yshift=-6.5pt,xshift=3pt]{*} ++(24pt,0);        
        
    \end{axis}        
\end{tikzpicture}
\caption{Results obtained for Predictability for simulated (top) and real-world (bottom) user studies. Results are shown grouped by gaze behavior (left) and navigation scenario (right). Significant paired differences are indicated with ($*$) for $p<0.05$, and with ($**$) for $p<0.001$.}
\label{fig:pred}
\end{figure*}

Comparing gaze behaviors within each navigation scenario (Fig.~\ref{fig:pred}a), the only significant results is that for the ENT scenario, the POS gaze is significantly more predictable compared to the D gaze. This is in opposition to Hypothesis 2, which expected the human-oriented gaze to be the least predictable when it gazes in a direction it is not travelling in. The ENT scenario captures this situation, while the other two do not. Therefore, it is surprising that the POS was perceived as the most predictable in this scenario.

Comparing navigation scenarios given each gaze behavior (Fig.~\ref{fig:pred}b), we see a common trend where the robot is the least predictable in the ENT scenario. However, this effect is not significant for the POS gaze. This supports Hypothesis 2 for the POL gaze, but not for the POS gaze.

\subsubsection{Real-world studies}
There was significant interaction between the gaze behavior and the navigation scenario for the Co-Presence metric ($F(6,210)=2.780$, $p<0.05$), therefore, post hoc paired sample $t$-test was performed for each gaze-scenario pair. 

Comparing gaze behaviors within each navigation scenario (Fig.~\ref{fig:pred}c), we only significant result is that for the EXT scenario, the D gaze is significantly less predictable compared to all other gazes.

Comparing navigation scenarios given each gaze behavior (Fig.~\ref{fig:pred}d), a similar observation is made -- for the D gaze, the EXT scenario is significantly less predictable compared to all other scenarios. This suggests that the combination of D gaze and EXT scenario results is significantly less predictable compared to all other situations. As person-oriented gazes in the ENT scenario are perceived as equally as predictable compared to other gazes and scenarios, Hypothesis 2 is not supported.

\subsection{Perceived Safety}
\subsubsection{Simulated studies}
There was significant interaction between the gaze behavior and the navigation scenario for the Safety metric ($F(6,1404)=9.979$, $p<0.05$), therefore, post hoc paired sample $t$-test was performed for each gaze-scenario pair. 

\begin{figure*}[t]
\centering
\begin{tikzpicture}

    \definecolor{a}{RGB}{217,217,217}
    \definecolor{b}{RGB}{255,255,179}
    \definecolor{c}{RGB}{253,180,98}
    \definecolor{d}{RGB}{251,128,114}

    \pgfplotsset{
      /pgfplots/bar  cycle  list/.style={/pgfplots/cycle  list={%
            {a!67!black,fill=a,mark=none},%
            {b!67!black,fill=b,mark=none},%
            {c!67!black,fill=c,mark=none},%
            {d!67!black,fill=d,mark=none},%
         }
      },
    }

    \begin{axis}[
        name=mainplot,
        ybar,
        enlarge x limits=0.25,
        ylabel={Safe. (Simulated)},
        symbolic x coords={TW,ENT,EXT},
        xtick=data,
        xticklabels={,,},
        bar width=10pt,
        xtick pos=left,
        width = 0.55\columnwidth,
        height = 0.3\columnwidth,
        ymin = 1, ymax = 3.25,
        title=\small(a),
        title style={yshift=-1.25ex,},
    ]
    
    \addplot+ [
        error bars/.cd,
            y dir=both, y explicit,
    ] coordinates {
        (TW,2.5660)  +- (0, 0.03703)
        (ENT,2.4043) +- (0, 0.04139)
        (EXT,2.3319) +- (0, 0.04506)
    };

    \addplot+ [
        error bars/.cd,
            y dir=both, y explicit,
    ] coordinates {
        (TW,2.4298)  +- (0, 0.04117)
        (ENT,2.2383) +- (0, 0.04311)
        (EXT,2.5702) +- (0, 0.03700)
    };
    
    \addplot+ [
        error bars/.cd,
            y dir=both, y explicit,
    ] coordinates {
        (TW,2.4809)  +- (0, 0.04140)
        (ENT,2.3191) +- (0, 0.04403)
        (EXT,2.5915) +- (0, 0.03825)
    };    
    
    \addplot+ [
        error bars/.cd,
            y dir=both, y explicit,
    ] coordinates {
        (TW,2.3489)  +- (0, 0.04610)
        (ENT,2.4596) +- (0, 0.04090)
        (EXT,2.5106) +- (0, 0.03823)
    };
    
    \draw [arrows={Bar[left]-Bar[right,width=20pt]}] 
        (axis cs:TW,2.75) ++(-18pt, 0) -- node[midway, above, yshift=-6.5pt]{*} ++(36pt,0);
        
    \draw [arrows={Bar[left]-Bar[right,width=17.5pt]}] 
        (axis cs:ENT,2.6) ++(-18pt, 0) -- node[midway, above, yshift=-6.5pt]{*} ++(12pt,0);
        
    \draw [arrows={Bar[left]-Bar[right,width=12.5pt]}] 
        (axis cs:ENT,2.75) ++(-6pt, 0) -- node[midway, above, yshift=-6.5pt]{**} ++(24pt,0);        
    
    \draw [arrows={Bar[left,width=25pt]-Bar[right,width=7.5pt,length=12pt]}] 
        (axis cs:EXT,2.8) ++(-18pt, 0) -- node[midway, above, yshift=-6.5pt,xshift=-3pt]{**} ++(24pt,0);
        
    \draw [arrows={Bar[left,width=12.5pt]-Bar[right,width=30pt]}] 
        (axis cs:EXT,3.05) ++(-18pt, 0) -- node[midway, above, yshift=-6.5pt]{*} ++(36pt,0);   
    
    \end{axis}
    \definecolor{a}{RGB}{204,235,197}
    \definecolor{b}{RGB}{128,177,211}
    \definecolor{c}{RGB}{188,128,189}
    \begin{axis}[
        name=secondplot,
        at={(mainplot.north east)},
        xshift=0.5cm,
        anchor=north west,    
        ybar,
        enlarge x limits=0.175,
        symbolic x coords={D,PO,POS,POL},
        xtick=data,
        yticklabels={,,},
        xticklabels={,,},
        bar width=10pt,
        xtick pos=left,
        width = 0.55\columnwidth,
        height = 0.3\columnwidth,
        ymin = 1, ymax = 3.25,
        title=\small(b),
        title style={yshift=-1.25ex,},
    ]
    
    \addplot+ [
        error bars/.cd,
            y dir=both, y explicit,
    ] coordinates {
        (D,2.5660)  +- (0, 0.03703)
        (PO,2.4298)  +- (0, 0.04117)
        (POS,2.4809)  +- (0, 0.04140)
        (POL,2.3489)  +- (0, 0.04610)
    };

    \addplot+ [
        error bars/.cd,
            y dir=both, y explicit,
    ] coordinates {
        (D,2.4043) +- (0, 0.04139)
        (PO,2.2383) +- (0, 0.04311)
        (POS,2.3191) +- (0, 0.04403)
        (POL,2.4596) +- (0, 0.04090)
    };
    
    \addplot+ [
        error bars/.cd,
            y dir=both, y explicit,
    ] coordinates {
        (D,2.3319) +- (0, 0.04506)
        (PO,2.5702) +- (0, 0.03700)
        (POS,2.5915) +- (0, 0.03825)
        (POL,2.5106) +- (0, 0.03823)
    };
    
    \draw [arrows={Bar[left]-Bar[right,width=15pt]}] 
        (axis cs:D,2.725) ++(-12pt, 0) -- node[midway, above, yshift=-6.5pt]{*} ++(12pt,0);

    \draw [arrows={Bar[left,width=12.5]-Bar[right,width=37.5pt]}] 
        (axis cs:D,3.0) ++(-12pt, 0) -- node[midway, above, yshift=-6.5pt]{**} ++(24pt,0);
        
    \draw [arrows={Bar[left]-Bar[right,width=17.5pt]}] 
        (axis cs:PO,2.6) ++(-12pt, 0) -- node[midway, above, yshift=-6.5pt]{*} ++(12pt,0);

    \draw [arrows={Bar[left,width=22.5]-Bar[right,width=12.5pt]}] 
        (axis cs:PO,3.0) ++(-12pt, 0) -- node[midway, above, yshift=-6.5pt]{*} ++(24pt,0);        

    \draw [arrows={Bar[left]-Bar[right]}] 
        (axis cs:PO,2.725) ++(0pt, 0) -- node[midway, above, yshift=-6.5pt]{**} ++(12pt,0);

    \draw [arrows={Bar[left]-Bar[right,width=15pt]}] 
        (axis cs:POS,2.65) ++(-12pt, 0) -- node[midway, above, yshift=-6.5pt]{*} ++(12pt,0);

    \draw [arrows={Bar[left]-Bar[right]}] 
        (axis cs:POS,2.775) ++(0pt, 0) -- node[midway, above, yshift=-6.5pt]{*} ++(12pt,0);

    \draw [arrows={Bar[left,width=15]-Bar[right]}] 
        (axis cs:POL,2.675) ++(-12pt, 0) -- node[midway, above, yshift=-6.5pt]{*} ++(24pt,0);   
        
    \end{axis}
    \definecolor{a}{RGB}{217,217,217}
    \definecolor{b}{RGB}{255,255,179}
    \definecolor{c}{RGB}{253,180,98}
    \definecolor{d}{RGB}{251,128,114}
    \begin{axis}[
        at={(mainplot.below south west)},
        yshift=-0.35cm,
        anchor=north west,
        ybar,
        enlarge x limits=0.25,
        legend style={at={(0.5,-0.25)},
        anchor=north,legend columns=-1,
        /tikz/every even column/.append style={column sep=0.5cm}},
        ylabel={Safe. (Real-world)},
        symbolic x coords={TW,ENT,EXT},
        xtick=data,
        bar width=10pt,
        xtick pos=left,
        width = 0.55\columnwidth,
        height = 0.3\columnwidth,
        ymin = 1, ymax = 5.49,
        legend image code/.code={\draw [#1] (0cm,-0.075cm) rectangle (0.2cm,0.125cm); },
        title=\small(c),
        title style={yshift=-1.25ex,},
    ]
    
    \addplot+ [
        error bars/.cd,
            y dir=both, y explicit,
    ] coordinates {
        (TW,3.6944)  +- (0, 0.1370)
        (ENT,3.6389) +- (0, 0.1205)
        (EXT,3.2500) +- (0, 0.1661)
    };

    \addplot+ [
        error bars/.cd,
            y dir=both, y explicit,
    ] coordinates {
        (TW,3.6389)  +- (0, 0.1601)
        (ENT,3.8611) +- (0, 0.1601)
        (EXT,3.4444) +- (0, 0.1711)
    };
    
    \addplot+ [
        error bars/.cd,
            y dir=both, y explicit,
    ] coordinates {
        (TW,4.0000)  +- (0, 0.1491)
        (ENT,3.9722) +- (0, 0.1463)
        (EXT,4.0278) +- (0, 0.1090)
    };    
    
    \addplot+ [
        error bars/.cd,
            y dir=both, y explicit,
    ] coordinates {
        (TW,4.0000)  +- (0, 0.1737)
        (ENT,3.8333) +- (0, 0.1517)
        (EXT,3.9444) +- (0, 0.1433)
    };

    \draw [arrows={Bar[left,width=15pt]-Bar[right]}] 
        (axis cs:TW,4.35) ++(-6pt, 0) -- node[midway, above, yshift=-5pt]{*} ++(12pt,0);
        
    \draw [arrows={Bar[left,width=30pt]-Bar[right,width=10pt,length=12pt]}] 
        (axis cs:TW,4.9) ++(-18pt, 0) -- node[midway, above, yshift=-5pt,xshift=-3pt]{**} ++(36pt,0);
        
    \draw [arrows={Bar[left]-Bar[right]}] 
        (axis cs:ENT,4.35) ++(-6pt, 0) -- node[midway, above, yshift=-5pt]{*} ++(12pt,0);
        
    \draw [arrows={Bar[left,width=32.5pt]-Bar[right,width=10pt,length=12pt]}] 
        (axis cs:ENT,4.9) ++(-18pt, 0) -- node[midway, above, yshift=-5pt,xshift=-3pt]{**} ++(36pt,0); 
        
    \draw [arrows={Bar[left,width=20pt]-Bar[right]}] 
        (axis cs:EXT,4.4) ++(-6pt, 0) -- node[midway, above, yshift=-5pt]{*} ++(12pt,0);        
        
    \draw [arrows={Bar[left,width=45pt]-Bar[right,width=10pt,length=12pt]}] 
        (axis cs:EXT,5.0) ++(-18pt, 0) -- node[midway, above, yshift=-5pt,xshift=-3pt]{**} ++(36pt,0);

    \legend{D,PO,POS,POL}
    
    \end{axis}
    \definecolor{a}{RGB}{204,235,197}
    \definecolor{b}{RGB}{128,177,211}
    \definecolor{c}{RGB}{188,128,189}
    \begin{axis}[
        at={(secondplot.below south west)},
        yshift=-0.35cm,
        anchor=north west,  
        ybar,
        enlarge x limits=0.175,
        legend style={at={(0.5,-0.25)},
        anchor=north,legend columns=-1,
        /tikz/every even column/.append style={column sep=0.5cm}},
        symbolic x coords={D,PO,POS,POL},
        xtick=data,
        yticklabels={,,},
        bar width=10pt,
        xtick pos=left,
        width = 0.55\columnwidth,
        height = 0.3\columnwidth,
        ymin = 1, ymax = 5.49,
        legend image code/.code={\draw [#1] (0cm,-0.075cm) rectangle (0.2cm,0.125cm); },
        title=\small(d),
        title style={yshift=-1.25ex,},
    ]
    
    \addplot+ [
        error bars/.cd,
            y dir=both, y explicit,
    ] coordinates {
        (D,3.6944)  +- (0, 0.1370)
        (PO,3.6389)  +- (0, 0.1601)
        (POS,4.0000)  +- (0, 0.1491)
        (POL,4.0000)  +- (0, 0.1737)
    };

    \addplot+ [
        error bars/.cd,
            y dir=both, y explicit,
    ] coordinates {
        (D,3.6389) +- (0, 0.1205)
        (PO,3.8611) +- (0, 0.1601)
        (POS,3.9722) +- (0, 0.1463)
        (POL,3.8333) +- (0, 0.1517)
    };
    
    \addplot+ [
        error bars/.cd,
            y dir=both, y explicit,
    ] coordinates {
        (D,3.2500) +- (0, 0.1661)
        (PO,3.4444) +- (0, 0.1711)
        (POS,4.0278) +- (0, 0.1090)
        (POL,3.9444) +- (0, 0.1433)
    };
    
    \legend{TW,ENT,EXT}
    
    \end{axis}        
\end{tikzpicture}
\caption{Results obtained for Safety for simulated (top) and real-world (bottom) user studies. Results are shown grouped by gaze behavior (left) and navigation scenario (right). Significant paired differences are indicated with ($*$) for $p<0.05$, and with ($**$) for $p<0.001$.}
\label{fig:safe}
\end{figure*}
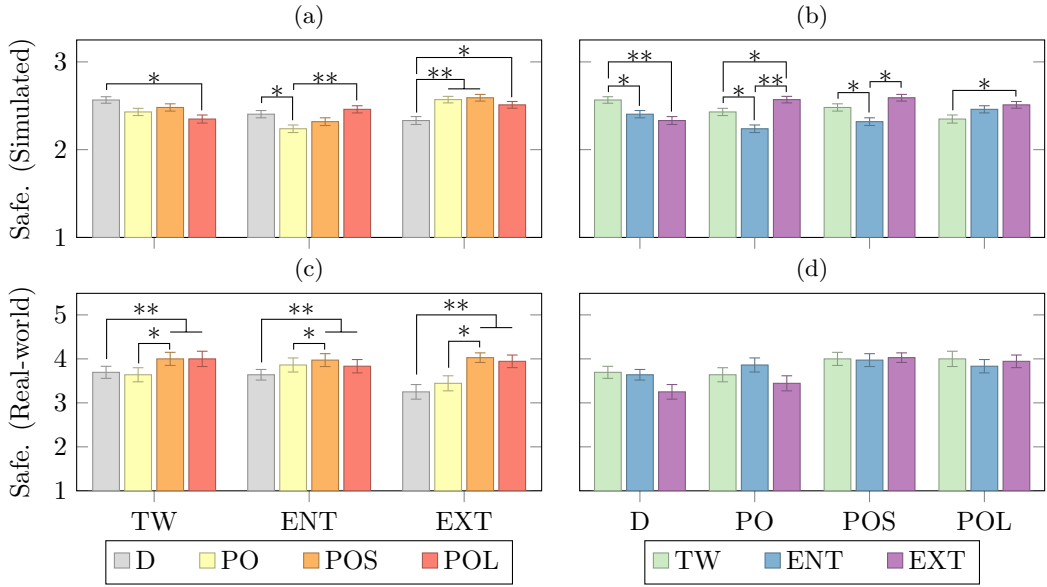

Comparing gaze behaviors within each navigation scenario (Fig.~\ref{fig:safe}a), we observe mixed results for each scenario. For the TW scenario, the D gaze is found to be significantly safer compared to the POL gaze. For the ENT scenario, both D and POL gazes are found to be significantly safer compared to the PO gaze. In contrast, for the EXT scenario, the D gaze is significantly less safe compared to all other gazes. Overall, Hypothesis 3 is only partially supported for the ENT and EXT scenarios.

Comparing navigation scenarios given each gaze behavior (Fig.~\ref{fig:safe}b), mixed observations are also made for each gaze. For the D gaze, the TW scenario is perceived as the safest. For all other gazes however, the EXT scenario is perceived as the safest. The ENT is perceived as the least safe for PO and POS gazes, while TW is the least safe for the POL gaze.

\subsubsection{Real-world studies}
Unlike the simulated studies, no significant interaction effects were found between gaze behavior and navigation scenario ($F(6,210)=1.842$, $p=0.09$). There were significant main effects exhibited by gaze behaviors ($F(3,105)=13.043$, $p<0.05$), but not by navigational scenarios ($F(2,70)=1.812$, $p=0.17$). Therefore post hoc paired sample $t$-tests were performed for investigate the impact of gaze behaviors after collapsing results in the scenario variable.


Comparing between gaze behaviors (Fig.~\ref{fig:safe}c), both person-oriented gaze behaviors (POS and POL) are perceived to be significantly safer compared to D and PO gazes, fully supporting Hypothesis 3.


\subsection{Naturalness}
\subsubsection{Simulated studies}
There was significant interaction between the gaze behavior and the navigation scenario for the Naturalness metric ($F(6,1404)=5.481$, $p<0.05$), therefore, post hoc paired sample $t$-test was performed for each gaze-scenario pair. 

Comparing gaze behaviors within each navigation scenario (Fig.~\ref{fig:pmu}a), for the TW scenario, the D gaze is perceived as significantly more natural compared to the POL gaze, in contradiction to Hypothesis 3. In contrast, Hypothesis 3 is supported for the other two scenarios, where the D gaze is perceived as the least natural gaze, while the POS gaze is the most natural.

Comparing navigation scenarios given each gaze behavior (Fig.~\ref{fig:pmu}b), both the D and PO gazes are perceived as the most natural when performing the TW scenario. There is no significant difference found between navigation scenarios for the two person-oriented gaze behaviors (POS and POL).

\subsubsection{Real-world studies}
Significant interaction effects were found between gaze behavior and navigation scenario ($F(6,210)=2.687$, $p<0.05$), therefore post hoc paired sample $t$-tests were performed for each gaze-scenario pair. 

Comparing gaze behaviors within each navigation scenario (Fig.~\ref{fig:pmu}c), both person-oriented gaze behaviors (POS and POL) are perceived to be significantly more natural compared to D and PO gazes for the EXT scenario, supporting Hypothesis 3. No significant differences are found between any of the other scenarios.

Comparing navigation scenarios given each gaze behavior (Fig.~\ref{fig:pmu}d), no significant differences are found for any gaze.

\section{Discussion}

\subsection{Simulated studies}
The results for the Social Presence Indicators are all generally in support for Hypothesis 1. The only contradiction to this is the results for the PMU metric, which suggest that the robot's intentions are less clear when it performs a short human-oriented gaze cue in the TW scenario. This may be due to the scenario itself being simple, where an undistracted robot focusing on moving forwards down the hallway is easiest to understand the intent of. However, other scenarios require some communication with the human to indicate what it will do. Another interesting observation is that the TW scenario has the lowest PBI for the path-oriented and person-oriented gazes. This is likely because the robot does not need to react to the human in this scenario, while there is always someone giving way to the other in the other scenarios, and therefore must be reacting to one's presence.




Hypothesis 2 is supported for the POL gaze. However, the POS gaze sees no difference in predictability between scenarios. This suggests that as long as the gaze is brief enough, people will not mistake the gaze intended to establish social awareness of the robot as an indication of where the robot will go. Interestingly however, the robot is also least predictable in the ENT scenario for the D and PO gazes, which might instead suggest there is some other underlying factor of the scenario which makes it inherently less predictable. In this case, the POS gaze has some inherent characteristic which is able to overcome the unpredictability of this scenario.

For Hypothesis 3, we have mixed results where the human-oriented gazes are perceived as safer and more natural than other gaze behaviors only in situations where the human and robot cross paths (ENT and EXT). This suggests that gaze cues towards human should be used for the human to feel acknowledged and safe when an interaction between the two is required, but may otherwise be distracting or off-putting in other scenarios.



\subsection{Real-world studies}

For the Social Presence Indicators, a similar but somewhat stronger trend to that found in the simulated study is shared in the real-world study. Hypothesis 1 is fully supported for the CP and PBI metrics. CP was also commented on the most by the participants, with 10 participants sharing comments similar to ``\textit{The eye contact was essential to me, knowing the robot acknowledged that I was there}''. 

Regarding PMU, the results suggest that for the ENT scenario, the PO gaze conveys the robot's intentions most clearly, while the for EXT scenario the human-oriented gazes were the clearest. Similar to the reasoning behind Hypothesis 2, the first observation might be due to how only the PO gaze looked at where it would travel in the ENT scenario. However for the second observation, the human is conveniently in the direction that the robot wants to travel towards in the EXT scenario and therefore the human-oriented gazes also adopted a pseudo-path-oriented gaze effect. Some participants commented that they would have preferred a combination between path-oriented and person-oriented gazes, where the robot looks at where it will travel, quickly gazes at the human, then returns back to looking at its path. This would allow the human-oriented gazes to be more predictable for all navigation scenarios. However, not every participant understood the intended meaning of the path-oriented gaze, and often attributed it to the robot being ``\textit{rude}'' or ``\textit{depressed}'', and ``\textit{[was not] sure what the robot was doing}''. A path-oriented gaze behavior that does not look at the ground may therefore help to convey the robot's intentions more clearly.


Hypothesis 2 is not supported from the real-world experiments. Together with the results regarding PMU, these results suggest that while path-oriented gazes help to convey the intention of the robot, it does not help to convey how the robot will perform its intended action. The general lack of differences between the predictability among the different gazes and scenarios may be due to how participants were shown and described to what the robot would do in each scenario. Two participants commented that ``\textit{it [was] a bit hard to [rate predictability] since I already knew each scenario}". Therefore, an improved experimental design may be required to verify our real-world results. 

The real-world results support Hypothesis 3. In all three navigational scenarios, the person-oriented gazes are perceived as safer compared to non-person-oriented gazes, with a marginal preference for shorter gazes.
Somewhat in contrast to the simulated results which suggest short person-oriented gazes feel more natural, both short and long person-oriented gazes felt equally more natural compared to the other gazes, but only for the EXT scenario. We received mixed opinions about which human-oriented gaze participants preferred. Some participants shared sentiments similar to ``\textit{I preferred it when the robot looked at me for a shorter period. The longer period felt like it was glaring at me}'', while other participants commented that ``\textit{The short period felt like a glitch. The longer gaze felt better}''. This suggests that an optimal gaze behavior should take user preference into account.

\subsection{Design Implications}
Overall, all three hypotheses are supported to an extent. However, whether our hypotheses are supported or not can depend on the navigational scenario. The simulated studies seem to suggest that for simple navigational scenarios such as the TW scenario where the robot and human don't have a significant interaction, person-oriented gazes may actually be detrimental to PMU, perceived safety, and naturalness of the robot compared to a simple default gaze behavior, despite exhibiting stronger co-presence. For the other scenarios, the person-oriented gazes tend to have a positive effect on all metrics. This suggests that the use of person-oriented gazes should be dependent on the scenario, and should be used when there is an important interaction, such as giving way, that will occur between the human and the robot.

Real-world experiments suggest that person-oriented gazes are equal to, or superior in the majority of metrics for all scenarios. Only the PMU metric deviated from this pattern for the ENT scenario. To improve the PMU for these gazes, particularly when the human is not in the direction of the path it will take, a PO behavior should be adopted before and after it gazes at the human. There were generally no significant differences in preference between the length of the gaze, although we believe this is due to differences in preferences between participants rather than the gazes themselves being negligibly different. Future work should be performed to examine if it is possible to determine user preferences in gaze periods. 

The results also show that the real-world results generally have much more muted differences between navigation scenarios compared to the simulated results. This may be due to differences in sample size or an artifact of using differently sized Likert scales between the simulated and real-world surveys.

The EXT scenario tended to draw out the most significant differences between gazes, particularly for the real-world results. We speculate that this is due to the EXT scenario being the most stimulating experience, as the robot is initially unseen by the human before it comes out of the room. Therefore, it is the situation which necessitates a cue towards the human the most. This suggests that robot cues to the human are more important for scenarios where the likelihood of a physical conflict is more likely.

\section{Limitations and Conclusion}
In this paper, we investigate the effects of varying the gaze behavior of a mobile robot for various common hallway navigation scenarios. Our work addresses conflicting results from previous studies and establishes that the beneficial use of robot gaze is scenario-dependent.

To answer our research questions, we performed simulated and real-world user studies across three different navigation scenarios. Our results vary based on which scenario is analysed, showing how people's perception of robotic gaze cues is situation-dependent. Overall, we find that scenarios with more potential conflicts benefit from person-oriented robot gaze cues the most, typically scoring higher in the Social Presence Inventory metrics, predictability, safety, and naturalness. However, the use of these gaze cues in simple scenarios requiring no interaction between the parties may yield no benefit, or as the simulated results suggest may even be detrimental to how socially present people perceive the robot to be.

While we have addressed multiple common hallway scenarios, which was a limitation of past studies, further investigation is required for understanding how the scenario should affect what kind of gaze behaviors should the robot employ. Our results may not generalize to wide variety of dynamic navigation scenarios likely to occur in the real world. For example, scenarios with two or more robots/persons introduces an extra layer of complexity to the system. The results from this study should provide valuable insight in developing systems that can tackle more complex navigation scenarios.

We have identified possible avenues for future work, summarized here. Firstly, we believe that a combined path-oriented and person-oriented gaze behavior will be able to alleviate some of the issues the person-oriented gaze behavior had with regard to the clarity of the robot's intentions. Improvements to the path-oriented gaze, such as looking forwards instead of the ground, are also suggested. Secondly, based on our mixed findings about people's preferences towards longer or shorter gazes, a more in-depth investigation into people's preferences over the human-oriented gaze period length is recommended, with the aim to be able to reactively adapt the gaze period length to people's preferences.

\section{Acknowledgements}
We would like to thank our collaborators Matthias Beyrle and Jan Faber at the German Aerospace Center, DLR, for their expert advice and guidance on this project.

\textbf{Ethics Approval:} This study has been approved by the Monash University Human Research Ethics Committee (Application ID: 25549).

\textbf{Funding:} This project was supported by the Australian Research Council Discovery Projects Grant, Project ID: DP200102858.

\textbf{Data Availability:} The datasets generated during and/or analysed during the current study are available from the corresponding author on reasonable request.


\bibliographystyle{IEEEtran}
\bibliography{sn-bibliography}


\end{document}